\documentclass[%
 reprint,
 amsmath,amssymb,
 aps,
]{revtex4-2}

\usepackage{graphicx}% Include figure files
\usepackage{dcolumn}% Align table columns on decimal point
\usepackage{bm}% bold math
\usepackage{hyperref}
\usepackage{url}
\usepackage{amsmath,amsfonts,amssymb}
\usepackage{bbm}
\usepackage{graphicx}
\usepackage{physics}

\usepackage{tikz}
\usepackage{pgfplots}
\pgfplotsset{compat=newest}

\usepackage{cleveref}
\usepackage{comment}
\usepackage{xcolor}
\usepackage{wrapfig}
\usepackage{soul}
\usepackage{enumitem}
\usepackage{float}

\usepackage[utf8]{inputenc} % allow utf-8 input
\usepackage[T1]{fontenc}    % use 8-bit T1 fonts
\usepackage{hyperref}       % hyperlinks
\usepackage{url}            % simple URL typesetting
\usepackage{booktabs}       % professional-quality tables
\usepackage{amsfonts}       % blackboard math symbols
\usepackage{nicefrac}       % compact symbols for 1/2, etc.
\usepackage{microtype}      % microtypography
\usepackage{xcolor}         % colors

\usepackage[acronym, nomain, nonumberlist]{glossaries}
\makenoidxglossaries
\glsdisablehyper

\newacronym{rl}{RL}{Reinforcement Learning}
\newacronym{ode}{ODE}{Ordinary Differential Equation}
\newacronym{td}{TD}{Temporal Difference}

\def\sgn{\text{sgn}}
\newcommand{\bw}{\mathbf{w}}
\newcommand{\bx}{\mathbf{x}}
\newcommand{\reals}{\mathbb{R}}
\newcommand{\appropto}{\mathrel{\vcenter{
  \offinterlineskip\halign{\hfil$##$\cr
    \propto\cr\noalign{\kern2pt}\sim\cr\noalign{\kern-2pt}}}}}

\newif\ifrebuttal

\DeclareRobustCommand{\rebuttal}[1]{\ifrebuttal{\begingroup\textcolor{red}{#1}\endgroup}\else{#1}\fi}

\DeclareRobustCommand{\rebuttal}[1]{\ifrebuttal{\begingroup\textcolor{red}{#1}\endgroup}\else{#1}\fi}

\begin{document}

\preprint{APS/123-QED}

\title{The RL Perceptron: Generalisation Dynamics of Policy Learning in High Dimensions}% Force line breaks with \\

\author{Nishil Patel\textsuperscript{1,*}, Sebastian Lee\textsuperscript{1,2}, Stefano Sarao Mannelli\textsuperscript{1,4}, Sebastian Goldt\textsuperscript{3}, and Andrew Saxe\textsuperscript{1,4,5,*}}
\affiliation{Gatsby Computational Neuroscience Unit, UCL\textsuperscript{1}, Imperial College London\textsuperscript{2}, International School of Advanced Studies (SISSA), Trieste, Italy\textsuperscript{3}, Sainsbury Wellcome Centre, UCL\textsuperscript{4}, CIFAR Azrieli Global Scholar, CIFAR\textsuperscript{5}}%Lines break automatically or can be forced with \\%
 \email{ucabnp2@ucl.ac.uk}
 \email{a.saxe@ucl.ac.uk}

% \author{Nishil Patel\textsuperscript{1}, Sebastian Lee\textsuperscript{1,2}, Stefano Sarao Mannelli\textsuperscript{1}, Sebastian Goldt\textsuperscript{3}, Andrew Saxe\textsuperscript{1,4,5}}

% \affil{\textsuperscript{1}Gatsby Computational Neuroscience Unit, UCL}
% \affil{\textsuperscript{2}Imperial College London}
% \affil{\textsuperscript{3}International School of Advanced Studies (SISSA), Trieste, Italy}
% \affil{\textsuperscript{4}Sainsbury Wellcome Centre, UCL}
% \affil{\textsuperscript{5}{§CIFAR Azrieli Global Scholar, CIFAR}

\begin{abstract}
~\gls{rl} algorithms have transformed many domains of machine learning. To tackle real-world problems, RL often relies on neural networks to learn policies directly from pixels or other high-dimensional sensory input. By contrast, much theory of RL has focused on discrete state spaces or worst-case analysis, and fundamental questions remain about the dynamics of policy learning in high-dimensional settings. Here, we propose a solvable high-dimensional ~\gls{rl} model that can capture a variety of learning protocols, and derive its typical policy learning dynamics as a set of closed-form ordinary differential equations (ODEs). We derive optimal schedules for the learning rates and task difficulty -- analogous to annealing schemes and curricula during training in RL -- and show that the model exhibits rich behaviour, including delayed learning under sparse rewards; a variety of learning regimes depending on reward baselines; and a speed-accuracy trade-off driven by reward stringency. Experiments on variants of the Procgen game ``Bossfight'' and Arcade Learning Environment game ``Pong'' also show such a speed-accuracy trade-off in practice. Together, these results take a step towards closing the gap between theory and practice in high-dimensional RL. 
\end{abstract}

\maketitle

%\tableofcontents

\section{\label{sec:level1}Introduction}

Thanks to algorithmic and engineering advancements, reinforcement learning (RL) methods have achieved super-human performance in a variety of domains, for example in playing complex games like Go~\citep{silver2016mastering, mnih2015human}. Reinforcement learning involves an agent in an environment that takes actions based on the given state of an environment it is exploring, for example, an agent playing chess must decide which action to take based on the state of the board. The map from states to actions is called a ``policy''. The overarching goal of the agent is to learn a policy that will allow them to maximise some kind of total reward, for example, a reward given for taking an opponent's piece in a game of chess. 

In cases where the state and action spaces are discrete and small enough, policies can be represented by simple look-up tables. However, the curse of dimensionality limits this approach to low-dimensional problems. In today's applications, environments are complex and policies are learnt directly from high-dimensional inputs representing the state of the environment, using neural networks~\citep{10020015}.

While comprehensive theoretical results exist for tabular~\gls{rl}, our theoretical grasp of ~\gls{rl} for high-dimensional problems requiring neural networks to represent the policy %non-linear function approximation 
remains limited, despite its practical success. The lack of a clear notion of similarity between discrete states further means that tabular methods do not address the core question of generalisation: how are values and policies extended to unseen states and across seen states~\citep{kirk_survey_2023}? Consequently, much of this theoretical work is far from the current practice of~\gls{rl}, which increasingly relies on deep neural networks to approximate policies and other~\gls{rl} components like value functions. Moreover, while~\gls{rl} theory has often addressed ``worst-case'' performance and convergence behaviour, their \emph{typical} behaviour has received comparatively little attention (see further related work in~\cref{relatedwork}).

Meanwhile, there is a long tradition of neural network theory that employs tools from statistical mechanics to analyse learning and generalisation in high-dimensional settings with a focus on \emph{typical} behaviours, as is usually the case in statistical mechanics. This theory was first developed in the context of supervised learning, see Refs.~\cite{seung1992statistical, engel2001statistical, carleo2019machine, doi:10.1146/annurev-conmatphys-031119-050745, gabrie2023neural} for classical and recent reviews. More recently, this approach yielded new insights beyond vanilla supervised learning, for example in curriculum learning~\citep{saglietti2022analytical}, continual learning~\citep{asanuma2021statistical, lee2021continual, lee2022maslow}, few-shot learning~\citep{doi:10.1073/pnas.2200800119} and transfer learning \citep{lampinen2018analytic,dhifallah2021phase,gerace2022probing}. However, policy learning has not been analysed using statistical mechanics yet -- a gap we address here by studying the generalisation dynamics of a simple neural network trained on a ~\gls{rl} task.

Our \textbf{goal} is to develop a theory for the typical dynamics of policy learning. For example, we would like to explain how problem properties and algorithmic choices impact how quickly a model will learn, or how effectively it will generalise. In order to achieve this goal we work on an analysis of a perceptron adapted online policy learning update which can be considered as an analogue to the REINFORCE algorithm~\citep{sutton:nips12} (more on this in~\cref{sec:connection_reinforce}). REINFORCE is the simplest online ``policy gradient'' algorithm. Policy gradient methods, which, despite their simplicity, underpin much of modern reinforcement learning with deep neural networks~\citep{haarnoja2018softactorcriticoffpolicymaximum, barth-maron2018distributional, schulman2017proximalpolicyoptimizationalgorithms}; consequently, an understanding of online policy learning dynamics is beneficial for transferable insights to more complex policy-gradient methods, and as a starting point from which to analyse more complex methods. We contrast further our method to existing results in ~\cref{relatedwork}. 

\begin{figure*}
    \centering
    \includegraphics[width=0.9\textwidth]{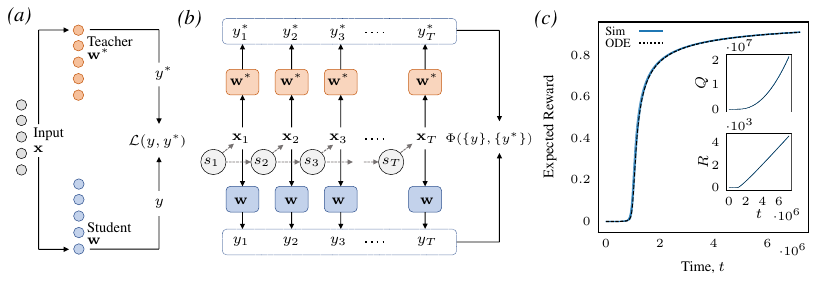}
    \caption{\textbf{The RL-Perceptron is a model for policy learning in high dimensions.} \emph{(a)} In the classic teacher-student model for supervised learning, a neural network called the student is trained on inputs~$x$ whose label $y^*$ is given by another neural network, called the teacher. \emph{(b)} In the RL setting the student moves through states $s_t$ making a series of $T$ choices given in response to inputs $x_t$. The RL-perceptron is an extension of the teacher-student model as we assume there is a `right' choice $y_t$ on each timestep given by a teacher network. The student receives a reward after $T$ decisions according to a criterion $\Phi$ that depends on the choices made and the corresponding correct choices.
    \emph{(c)} Example learning dynamics in the RL-perceptron for a problem with $T=12$ choices where the reward is given only if all the decisions are correct. The plot shows the expected reward of a student trained in the RL perceptron setting in simulations (solid) and for our theoretical results (dashed) obtained from solving the dynamical equations~\ref{eq:Rall} and~\ref{eq:Qall}. Finite size simulations and theory show good agreement. We reduce the stochastic evolution of the high dimensional student to the study of deterministic evolution of two scalar quantities $R$ and $Q$ (more details in Sec.~\ref{sec:ode}), their evolution are shown in the inset. \textit{Parameters: $D=900$, $\eta_1=1$, $\eta_2=0$, $T=12$.}\vspace*{-\baselineskip}}
    \label{fig1}
\end{figure*}

Our \textbf{approach} consists in considering a simple model of an ~\gls{rl} problem which we approach through the \textit{teacher-student} framework which allows for exact solutions, i.e.\ the derivation of equations describing the typical learning dynamics \textit{exactly}, we then analyse properties of the solution.
In the classic teacher-student model of supervised learning~\citep{gardner1989three, seung1992statistical}, a neural network called the student is trained on inputs $\bx$ whose labels $y^*$ are specified by another neural network called the teacher (see \cref{fig1}a). The goal of the student is to learn the function represented by the teacher from samples $(\bx, y^*)$. This framework enables an exact analysis, characterising generalisation error of learning algorithms over the entire learning trajectory. In many ~\gls{rl} settings, agents face sequential decision-making tasks that require a series of intermediate choices to successfully complete an episode. We map this process into the ~\gls{rl} perceptron, where the teacher can be viewed as specifying a `perfect policy network' that prescribes a reward signal used to train the student network representing the policy of the agent.
This setup lends itself to an analysis that exactly describes the \textit{average-case} dynamics over the entire learning trajectory. 

\subsection{Main results}

In this work, we develop a teacher-student framework for a high-dimensional sequential policy learning task, the \gls{rl} Perceptron and derive asymptotically exact \glspl{ode} that describe the typical learning dynamics of policy gradient~\gls{rl} agents in an online setting by building on classic work by~\citet{saad1995line}, see \cref{sec:ode}.

Using these \glspl{ode} we can characterize learning behaviour in various scenarios: We investigate sparse delayed reward schemes, the impact of negative rewards (\cref{sec:protocols}), derive optimal learning rate schedules and episode length curricula, and recover common annealing strategies (\cref{sec:scheduling}). We identify ranges of learning rates for which learning is `easy,' and `hybrid-hard' (\cref{sec:phasediagram}). We also identify a speed-accuracy trade-off driven by reward stringency (\cref{sec:speed-acc}).

In \cref{sec:experiments}, we demonstrate a similar speed-accuracy trade-off in simulations of high-dimensional policy learning from pixels using the procgen environment ``Bossfight''~\citep{cobbe2019procgen} and the Arcade Learning environment game ``Pong''~\citep{machado18arcade}.
A link to the code and instructions for running all simulations and experiments are given in in \cref{app:exper}.

\subsection{\label{relatedwork}Further related work}

\textbf{Sample complexity in RL.} \quad An important line of work in the theory of RL focuses on the sample complexity and other learnability measures for specific classes of models such as tabular~\gls{rl}~\citep{azar2017minimax, zhang2020almost}, state aggregation~\citep{dong2019provably}, various forms of Markov Decision Processes (MDPs)~\citep{jin2020provably, yang2019sample, modi2020sample, ayoub2020model, du2019provably, zhang2022efficient}, reactive Partially-Observable
Markov Decision Processes (POMDPs)~\citep{krishnamurthy2016pac}, and FLAMBE~\citep{agarwal2020flambe}. Here, we are instead concerned with the learning dynamics: how do reward rates, episode length, etc.\ influence the speed of learning and the final performance of the model.

\textbf{Statistical learning theory for RL} aims at finding complexity measures analogous to the Rademacher complexity or VC dimension from statistical learning theory for supervised learning~\cite{bartlett2002rademacher, vapnik2015uniform}. Proposals include the Bellman Rank~\cite{jiang2017contextual}, or the Eluder dimension~\citep{russo2013eluder} and its generalisations~\citep{jin2021bellman}. This approach focuses on worst-case analysis, which typically differs significantly from practice (at least in supervised learning~\citep{zhang2021understanding}). % Although the conceptual gap between value-based methods and policy gradient methods has been bridged recently~\citep{ghosh2020operator}, 
Furthermore, complexity measures for~\gls{rl}~are generally more suitable for value-based methods; policy gradient methods have received less attention despite their prevalence in practice~\cite{bhandari2019global, agarwal2021theory}. We focus instead on average-case dynamics of policy-gradient methods.
% ; in addition to the model class, many results are subject to further restrictions on the dimensionality of one parameter of the problem (e.g. state/action space), or properties of the model such as realisability and closedness.

% Results of the type discussed above are almost exclusively concerned with worst-case analysis, which can be far from the typical realisable case in practice~\cite{}. Our work on the other hand studies average-case performance, as well as the full dynamics of learning, by making use of the formalism and methods of statistical mechanics, which has been used to study various problems in deep learning~\cite{}. This provides a complementary perspective to the worst-case results that may in some instances be more insightful to the practitioner, and is arguably the key difference in our approach. Previous works to have incorporated ideas from statistical mechanics into analysis of~\gls{rl} include use of mean-field theory to study convergence of feature learning in temporal difference methods by~\citet{zhang2020can}; modelling of policy gradient learning dynamics as drift diffusion by~\citet{fabbricatore2022gradient}; and formulation of distributions over trajectories in MDPs as partition functions with their own Bellman operator by~\citet{rahme2019theoretical}.

\textbf{Dynamics of learning.} \quad A series of recent papers considered the dynamics of temporal-difference learning and policy gradient in the limit of wide two-layer neural networks~\citep{cai2019neural, zhang2020temporal, % SG: zhang2020temporal has mean-field in the title, but they consider a large parameter which is the \alpha of the lazy regime
agazzi2021global, agazzi2022temporal}. These works focus on one of two ``wide''
limits: the neural tangent kernel~\citep{jacot2018neural, du2019gradient} or
``lazy'' regime~\citep{chizat2019lazy}, where the network behaves like an
effective kernel machine and does not learn data-dependent features, which is
key for efficient generalisation in
high-dimensions. \citet{DBLP:journals/corr/abs-2102-13089} consider a framework
to study value learning using the temporal-difference algorithm, their finding
provides insights into quantities to track during learning to describe the
dynamics of representations, but they still rely on input from the RL problem in
consideration. In our setting, the success of the student crucially relies on
learning the weight vector of the teacher, which is hard for lazy
methods~\citep{ghorbani2019limitations, ghorbani2020neural, chizat2020implicit,
  refinetti2021classifying}. The other ``wide'' regime is the mean-field limit of interacting particles~\citep{mei2018mean, chizat2018global, rotskoff2018interacting}, where learning dynamics are captured by a non-linear partial differential equation. While this elegant description allows them to establish global convergence properties, the resulting equations are hard to solve in practice, and a further analysis is therefore difficult. The ODE description we derive here instead will allow us to describe a series of effects in the following sections. Closer to our work is \citet{bordelon2023loss}, who give a typical-case analysis of the dynamics of temporal-difference learning, which learns the value function rather than the policy as we do here. We will come back to this point in our Discussion. Finally, \citet{Rubin_2010} analysed the dynamics of the Tempotron \citep{gutig2006tempotron}, a neuron model that learns spike timing–based decisions, and similarly to our work they consider sparse rewards, however beyond this similarity the paper doesn't connect to RL and their update rules are substantially different.

\section{\label{setup_sec}The RL Perceptron: setup and learning algorithm}
The RL Perceptron considers a task, illustrated in~\cref{fig1}b, where an agent takes a sequence of choices over an episode (episodes are length $T$ i.e. a choice is made at every timestep $t \in \{1, \dots, T\}$). At each timestep, the agent occupies some state $s_t$ in the environment, and receives some high-dimensional observation $\bx_t \in \reals^D$ conditioned on $s_t$ with $t=1,\ldots,T$. In the preceding we consider the POMDP formalism (detailed in \ref{sec:MDP}), we could equally consider the MDP formalism (also detailed in \ref{sec:MDP}), where $D$ could be considered as the feature dimension. The student network determines the actions made at each $t$. We study the simplest possible student network, a perceptron with weight vector~$\bw \in \reals^D$ that takes in observations $\bx_t$ and outputs $y(\bx_t) = \sgn(\frac{\bw^\intercal \bx_t}{\sqrt{D}})$. We interpret the outputs $y_t$ as binary actions, for example, whether to go left or right in an environment. As the agent makes choices in response to the output of the student network with observations as inputs, the student is analogous to a \textit{policy} network; where the \textit{deterministic} policy $\pi_\bw(a_t\|\bx_t) = \frac{1}{2}(1 + a_t y_t)$ specifies the probability of taking action $a_t \in \{-1,1\}$ given observation $\bx_t$ (i.e the action taken $a_t$ is always equal to the action $y_t$ specified by the student). In \gls{rl}, a policy defines the learning agent's way of behaving at a given time. The teacher network $\mathbf{w^*} \in \reals^D$ can be considered as a `\textit{perfect} policy network', that specifies the `correct' decision ($y^*_t = \sgn(\frac{\mathbf{w^*}^\intercal \bx_t}{\sqrt{D}})$) to be made at every step (the policy which the student aims to emulate through the learning process). \rebuttal{A significant simplification for analytical tractability is that the actions do not influence state transitions (i.e the observations $\mathbf{x}_t$ are i.i.d.).} The crucial point is that the student does not have access to the correct choices, but only receives feedback at timestep $t$ in the form of some (non-Markovian) reward $R_t(y_{1:t}, y^*_{1:t}, \Phi)$ which is conditioned on all the actions taken ($y_{1:t}$) and `correct' actions ($y^*_{1:t}$) up to timestep $t$, and the condition for reward, $\Phi$. For instance, $\Phi$ could be the condition that the agent receives a reward on the $T$th step only if all choices in an episode are correctly made, and a penalty otherwise---a learning signal that is considerably less informative than in supervised learning. We will see in \cref{sec:phasediagram} that receiving penalties is not always beneficial.
In the model, $T$ plays the role of \textit{difficulty}, a more complex task may be defined by the number of correct decisions to take in order to receive a reward. An alternative notion of difficulty would be to consider a planted version of the convex perceptron used in jamming \citep{Franz_2016}. However, this formulation would not allow some manipulations --e.g. dense rewards described in \cref{sec:protocols}-- that are standard practice in \gls{rl}. 

To train the network (learn a policy), we consider a weight update where the student evolves after the $\mu$th episode as

 \begin{equation}
     \mathbf{w}^{(\mu+1)}  = \mathbf{w}^{(\mu)}+\frac{\eta}{T\sqrt{D}} \sum_{t=1}^{T}y_t^{(\mu)}  \mathbf{x}_{t}^{(\mu)}  G_t^{(\mu)},
 \label{eq:w_update}
 \end{equation}

where $G_t = \sum_{t^\prime = t}^T \gamma^{t^\prime-t}R_{t^\prime}$ is the total discounted reward from time $t$, $\gamma \in \left(0,1\right]$ is the discount factor, $\eta$ is the learning rate, and superscript $\mu$ denotes variables from the $\mu$th timestep in the algorithm. (N.B. algorithm time and episode time are different and are respectively denoted by $\mu$ and $t$). \rebuttal{The motivation for this update comes as a modification of a Hebbian update that: occurs after an episode that can contain multiple observations and actions, and is weighted by $G_t$ (which also can depend on previous observations and actions). We comment on the connection of this update to policy gradient methods in the next section.}
For the analysis and simulations in \ref{results} we consider $\gamma=1$ and largely restrict ourselves to receiving sparse reward, i.e. where reward/penalty is only received at the end of an episode upon successful/unsuccessful completion of the episode; in this setting, the total discounted reward may be written as
\begin{equation}
    G_t = r_1\mathbb{I}(\Phi(y_{1:T}, y^*_{1:T})) -r_2 (1-\mathbb{I}(\Phi(y_{1:T}, y^*_{1:T}))) \quad \forall t,
    \label{eq:discount}
\end{equation}

where $r_1$ is the reward, $r_2$ is the penalty, $\mathbb I$ is an indicator function and $\Phi$ is the boolean criterion that determines whether the episode was completed successfully---for instance, $\mathbb{I}(\Phi) = \prod_t^T\theta(y_ty^*_t)$ (where $\theta$ is the step function) if the student has to get every decision right in order to receive a reward. The reward/penalty can be amalgamated into $\eta$: we define $\eta_1=\eta r_1$ and $\eta_2=\eta r_2$, essentially `positive' and `negative' learning rates, and we use these instead of $r_1$ and $r_2$ in the remaining text.
 
Note that in the case of $T = 1$, $\eta_1 = 0$, $\eta_2 > 0$, and $\mathbb{I}(\Phi) =\theta(-y y^*)$, (i.e the learning rule updates the weight only if the student is \textit{incorrect} on a given sample) we recover the famous perceptron learning rule of supervised learning~\citep{rosenblatt1961neurodynamics}.

% We ground the MDP formalism of this setup: the observations $\bx$ correspond to the environment states $s$ themselves. The agent transitions between states at each timestep with independent samples $s_t \sim {N}(\mathbf{0}, \mathbf{\mathbbm{1}}_D)$ and $\bx_t = s_t$. There is another interpretation where we consider low dimensional latent states with $\bx$ as noisy observations of these states (the POMDP setup), we detail this interpretation in appendix~\cref{app_PG_update}. In this way, the teacher network is a part of the environment by specifying reward. 
\subsection{Connection to REINFORCE policy gradient}
\label{sec:connection_reinforce}
\rebuttal{ The update \cref{eq:w_update} can be related to the REINFORCE policy gradient algorithm~\citep{sutton:nips12}. REINFORCE is a Monte-Carlo policy gradient method. Policy gradient methods aim to optimise parameterised policies with respect to the \textit{return} $J$ (total expected reward over an episode). In the case of REINFORCE the return is estimated from episode samples - i.e. single episodes are sampled by acting under a current policy, and these sampled episodes are used to update said policy. In an arbitrary environment where an agent occupies states $s_t$ and may take actions $a_t$ by acting under some policy $\pi$ parameterised by $\theta$, the policy gradient is given by}
\begin{align}
\nabla_{\theta} J &=\left\langle\sum_{t=0}^{T-1} \nabla_{\theta} \log \pi_{\theta}\left(a_t | s_t\right)G_t\right\rangle,
    \label{eq: policy gradient}
\end{align}

\rebuttal{and the REINFORCE update of $\theta$ at the $\mu$th timestep in the algorithm for the $\mu$th sampled episode is hence given by}
\begin{align}
    \label{eq:reinforce_update}
\theta^{(\mu + 1)} &= \theta ^{(\mu)} + \eta \sum_{t=0}^{T-1} \nabla_{\theta} \log \pi_{\theta^\mu}\left(a_t^{(\mu)} | s_t^{(\mu)}\right)G_t^{(\mu)},
\end{align}
 
\rebuttal{ in the RL Perceptron setup the true policy is deterministic and given by $\pi_\bw(a_t |\bx_t) = \frac{1}{2}(1 + a_t\sgn(\frac{\bw^\intercal \bx_t}{\sqrt{D}}))$. In this case, \cref{eq:reinforce_update} becomes }

\begin{align}
    \mathbf{w}^{(\mu+1)}  = \mathbf{w}^{(\mu)} + \frac{\eta}{\sqrt{D}} \sum_{t=0}^{T-1} y_t^{(\mu)}\mathbf{x}_t^{(\mu)} G_t^{(\mu)} \delta ({\bw^{(\mu)}}^\intercal \bx^{(\mu)}_t) 
\end{align}

\begin{figure*}
    \centering
    \includegraphics[width=\textwidth]{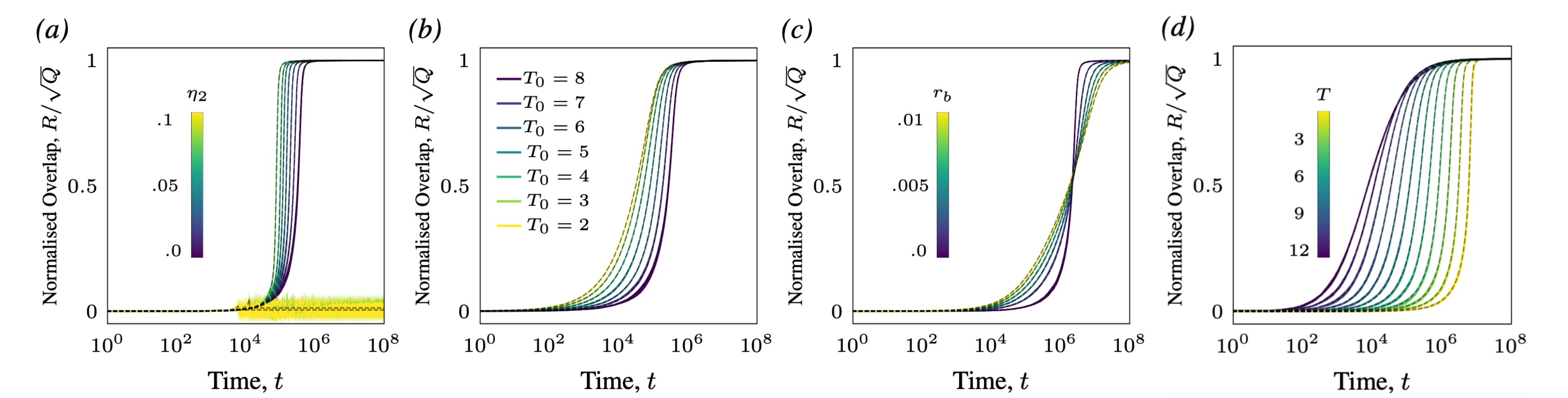}
    \caption{\textbf{ODEs accurately describe diverse learning protocols.} Evolution of the normalised student-teacher overlap $\rho$ for the numerical solution of the ODEs (dashed) and simulation (coloured) in three reward protocols. All students receive a reward of $\eta_1$ for getting all decisions in an episode correct, and additionally: \emph{(a)} A penalty $\eta_2$ (i.e. negative reward) is received if the agent does not survive until the end of an episode. \emph{(b)} An additional reward of 0.2 is received at timestep $T_0$ if the agent survives beyond $T_0$ timesteps. \emph{(c)} An additional reward $r_b$ is received at timestep $t$ for every correct decision $y_t$ made in an episode. \rebuttal{\emph{(d)} Episode length $T$ is varied.}\textit{Parameters: $D=900$, $T=11$, $\eta_1=1$.}\vspace*{-\baselineskip}}
    \label{fig2}
\end{figure*}

\rebuttal{where $\delta(\cdot)$ is the Dirac delta. We have inserted observations $\bx_t$ in place of states $s_t$ and we can replace $a_t$ by $y_t$ due to the deterministic policy. This is not a tractable update, but we can see the structural similarity to \cref{eq:w_update}, which can be fully recovered by replacing the gradient of $\text{sgn}(\bw^\intercal \bx)$ with the linearized gradient $\left(\nabla_{\bw} \bw^\intercal \bx \right)$ and rescaling by $T$. We would like to comment on policy gradients in general, so we verify that this linearization does not change the qualitative behaviour of the REINFORCE update in \cref{app_exact_reinforce} - where we plot the learning dynamics for a variety of reward settings when training under the exact REINFORCE policy gradient update and acting under a logistic policy. \Cref{app:reinforce_sims}a, c and d can be compared to \cref{fig2}a, b and c respectively, and \cref{fig2}d can be compared to \cref{fig5} for verification of consistent qualitative behaviour.}

\rebuttal{This connection of the RL Perceptron update to policy gradients is analogous to the connection of the classic Perceptron update to gradient descent; the Perceptron update rule in classic supervised learning is equal to the linearized version of the single sample gradient descent update for binary classification on square loss with a perceptron.}

\subsection{Connection to Markov decision processes}
\label{sec:MDP}
Formally, most RL problems can be described as a \textit{Markov Decision Process} (MDP) or a \textit{Partially-Observable Markov Decision Process} (POMDP), see \cref{app_PG_update}. Therefore, a lot of theoretical work on reinforcement learning has been formulated in this framework~\citep{10.5555/3312046}. To make it easier to connect the RL perceptron with this literature, we show in \cref{setup_sec} that the RL perceptron can be formulated as either an MDP or POMDP  with non-Markovian rewards. In a nutshell, the observations
$\mathbf{x_t}$ in the implementation described can be
thought of as high-dimensional states $s_t$  (MDP) or as the noisy
high-dimensional observations of underlying low-dimensional latent states $s_t$
(POMDP). Each interpretation naturally leads to different extensions. The MDP
framing is able to incoorporate kernelized high-dimensional feature maps of the
underlying state, and the POMDP framing is more amenable to tractable
calculation of expectations for trajectories involving state and action
dependent state transitions. We do not explore the connection further, as we are primarily interested in the dynamics of learning.

\section{Theoretical Results}%
\label{results}

The RL perceptron enables an analytical investigation of average dynamics through the identification and characterization of a few relevant order parameters, as explained in \cref{sec:ode}. This approach significantly simplifies the problem by transitioning from a high-dimensional to a low-dimensional framework. Moreover, it offers adaptability for characterizing and comparing various learning protocols, as detailed in \cref{sec:protocols}. On a practical level, the derived equations allow for the determination of optimal learning rate annealing strategies to maximize expected rewards, and the use of a curriculum protocol enhances training efficiency (\cref{sec:scheduling}). At a fundamental level, studying the low-dimensional equations provides valuable insights into the nature of the problem. Firstly, we observe in \cref{sec:phasediagram} that the presence of negative rewards can result in suboptimal fixed points and a counter-intuitive slowing down of dynamics near the emergence of such suboptimal fixed points. Secondly, in \cref{sec:speed-acc}, we demonstrate that several protocols aimed at expediting the initial learning phase actually lead to poorer long-term performance.

\subsection{A set of ODEs captures the learning dynamics of an RL perceptron exactly}%
\label{sec:ode}
\looseness=-1
The goal of the student during training is to emulate the teacher as closely as possible; or in other words, have a small number of disagreements with the teacher $y(\bx) \neq y^*(\bx)$. The generalisation error is given by the average number of disagreements
\begin{eqnarray}
    \label{eq:eg}
    \epsilon_g &&\equiv \frac{1}{2}\langle (y-y^*)^2 \rangle \\ 
    &&= \frac{1}{2}\left( 1- \left\langle \sgn \left(\frac{\bw^* \cdot \bx}{ \sqrt D}\right) \sgn \left(\frac{\bw \cdot \bx }{ \sqrt D}\right) \right\rangle \right)\nonumber \\ 
    &&= \frac{1}{2} \left (1 -  \left\langle \sgn \left(\nu \right) \sgn \left( \lambda \right) \right\rangle \right)
\end{eqnarray}
% \begin{eqnarray}
%     \label{eq:eg}
%     \epsilon_g \equiv \langle y(\bx) y^*(\bx) \rangle &&= \left\langle \sgn \left(\bw^* \cdot \bx / \sqrt D\right) \sgn \left(\bw \cdot \bx / \sqrt D\right) \right\rangle \nonumber \\ 
%     &&= \langle \sgn (\nu) \sgn (\lambda) \rangle
% \end{eqnarray}
where the average $\langle \cdot \rangle$ is taken over the inputs $\bx$, and we have introduced the scalar pre-activations for the student and the teacher, $\lambda \equiv \bw \cdot \bx / \sqrt D$ and  $\nu \equiv \bw^* \cdot \bx / \sqrt D$, respectively. We can therefore transform the high-dimensional average over the inputs $\bx$ into a low-dimensional average over the pre-activations $(\lambda, \nu)$. By specifying a distribution over observations, $\mathbf{x}_t \sim \mathcal{N}(\mathbf{0}, \mathbf{\mathbbm{1}}_D)$, the average in \ref{eq:eg} can be carried out by noting that the tuple $(\lambda, \nu)$ follow a jointly Gaussian distribution with means $\langle \lambda \rangle= \langle \nu \rangle= 0$ and covariances
\begin{eqnarray}
    \label{eq:order-parameters}
    Q \equiv \langle \lambda^2 \rangle = \frac{\bw \cdot \bw}{D}, \quad R \equiv \langle \lambda \nu \rangle = \frac{\bw \cdot \bw^*}{D} \nonumber \\
    \qq{and} S \equiv \langle \nu^2  \rangle = \frac{\bw^* \cdot \bw^*}{D}.
\end{eqnarray}
These covariances, or \textit{order-parameters} as they are known in statistical physics, have a simple interpretation. The overlap $S$ is simply the length of the weight vector of the teacher, for the sake of neatness of equations, we choose $S=1$. Likewise, the overlap $Q$ gives the length of the student weight vector; however, this is a quantity that will vary during training. For example, when starting from small initial weights, $Q$~will be small, and grow throughout training. Lastly, the ``alignment'' $R$ quantifies the correlation between the student and the teacher weight vector. At the beginning of training, $R \approx 0$, as both the teacher and the initial condition of the student are drawn at random. As the student starts learning, the overlap $R$ increases. Evaluating the Gaussian average in \ref{eq:eg} shows that the generalisation error is then a function of the normalised overlap $\rho = R / \sqrt Q$, and given by
\begin{equation}
    \label{eq:eg-explicit}
    \epsilon_g = \frac{1}{\pi}\arccos\left(\frac{R}{\sqrt Q}\right)
\end{equation}
The crucial point is that the description of the high-dimensional learning problem has been reduced from $D$ parameters to two time-evolving quantities, $Q$ and $R$, which are self-averaging in the $D\rightarrow \infty$ limit. We now discuss their dynamics.

\textbf{The dynamics of order parameters:} At any given point during training, the value of the order parameters determines the test error via \cref{eq:eg-explicit}. But how do the order parameters evolve during training with the stochastic update rule \cref{eq:w_update}? We followed the approach of \citet{kinzel1990improving, saad1995line, biehl1995learning} to derive a set of dynamical equations that describe the dynamics of the student in the thermodynamic limit where the input dimension goes to infinity. The general ODEs derived in~\cref{app:derivations} are given below:
\begin{eqnarray}
     \frac{dR}{d\alpha} &&= \frac{\eta}{T}\left\langle \sum_{t=1}^{T} \text{sgn}(\lambda_t) \nu_t  G_t\right\rangle \label{eq:Rgeneral} \\
     \frac{dQ}{d\alpha} &&= \frac{2\eta}{T}\left\langle \sum_{t=1}^{T} \text{sgn}(\lambda_t) \lambda_t  G_t\right\rangle + \frac{\eta^2}{T^2} \left\langle \sum_{t=1}^{T} G_t^2\right\rangle \label{eq:Qgeneral}
\end{eqnarray}
where $\alpha$ serves as a continuous time variable in the limit $D\to\infty$ (not to be confused with $t$ which counts episode steps). In this way the stochastic evolution of the student in high-dimension has been mapped to the deterministic evolution of two order-parameters in a continuous time description.

We give explicit dynamics for a variety of learning protocols (different protocols are encapsulated by the functional form of the discounted reward, $G_t$). Due to the length of these expressions, we report the explicit ODEs of the dynamics in the supplementary material in \cref{app:derivations}. In Secs.~\ref{sec:scheduling},\ref{sec:phasediagram},\ref{sec:speed-acc} we devote analysis to the reward condition where the agent must survive until the end of an episode to receive reward and receives a penalty otherwise (this is described by the $G_t$ given in \ref{eq:discount} and with $\mathbb{I}(\Phi) = \prod_t^T\theta(y_ty^*_t)$), as such, we explicitly state the ODEs for the order parameters in this below:
\begin{eqnarray}
    \label{eq:Rall}
     \frac{dR}{d\alpha} &&= \frac{\eta_1 + \eta_2}{\sqrt{2\pi}}\left (1 + \frac{R}{\sqrt{Q}} \right)P^{T-1} - \eta_2 R \sqrt{\frac{2}{\pi Q}}\\
     \frac{dQ}{d\alpha} &&= (\eta_1 + \eta_2)\sqrt{\frac{2Q}{\pi}}\left (1 + \frac{R}{\sqrt{Q}} \right) P^{T-1} \nonumber \\
     \label{eq:Qall}
     &&\qquad - 2\eta_2\sqrt{\frac{2Q}{\pi}} +  \frac{(\eta_1^2 - \eta_2^2)}{T}P^T + \frac{\eta_2^2}{T}, 
\end{eqnarray}
where $P = 1- \text{cos}^{-1}(R/\sqrt{Q})/\pi $ is the probability of a single correct decision. While our derivation of the equations follow heuristics from statistical physics, we anticipate that their asymptotic correctness in the limit $D\to\infty$ can be established rigorously using the techniques of \citet{goldt2019dynamics, veiga2022phase, arnaboldi2023high}. We illustrate the accuracy of these equations already in finite dimensions ($D=900$) in~\cref{fig1}c, where we show the expected reward, as well as the overlaps $R$ and $Q$, of a student as measured during a simulation and from integration of the dynamical equations (solid and dotted lines, respectively).

The derivation of the dynamical equations that govern the learning dynamics of the RL perceptron are our first main result. Equipped with this tool, we now analyse several phenomena exhibited by the RL perceptron through a detailed study of these equations.

\subsection{Learning protocols}%
\label{sec:protocols}

\begin{figure*}
    \centering
    \includegraphics[width=0.9\textwidth]{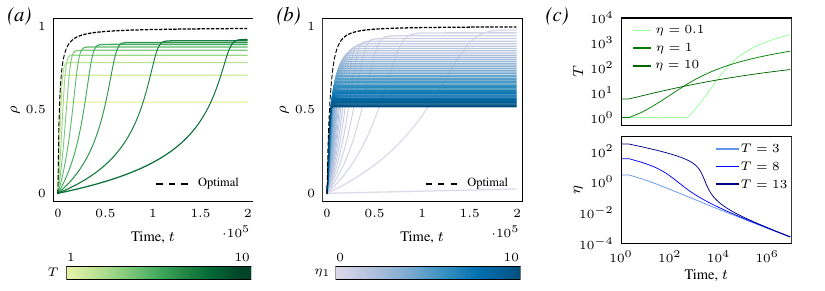}
    \caption{\textbf{Optimal schedules for episode length $T$ and learning rate $\eta$}. \emph{(a)} Evolution of the normalised overlap under optimal episode length scheduling (dashed) and various constant episode lengths (green). \emph{(b)} Evolution of the normalised overlap under optimal learning rate scheduling (dashed) and various constant learning rates (blue). \emph{(c)} Evolution of optimal $T$ (green) and $\eta$ (blue) over learning. \textit{Parameters: $D=900$, $Q=1$, $\eta_2=0$, (a) $\eta_1=1$, (b) $T=8$.}\vspace*{-\baselineskip}}
    \label{fig3}
\end{figure*}

The RL perceptron allows for the characterization of different RL protocols by adapting the reward condition $\Phi$. We considered the following three settings: % did mention this two times already at this point , for instance in the most stringent case $\left(\mathbb{I}(\Phi) = \prod_t^T(y_ty^*_t)\right)$ which requires the agent to `survive' till the end of an episode to receive a reward. 

\textbf{Vanilla:} The dynamics in the case without penalty and where survival of entire episode is required for reward ($G_t = \prod_t^T\theta(y_ty^*_t)$), is shown in~\cref{fig2}d. Rewards are sparsest in this protocol, as a result we observe a characteristic initial plateau in expected reward followed by a rapid jump. The length of this plateau increases with $T$, consistent with the notion that sparser reward slow learning~\citep{vasan2024revisitingsparserewardsgoalreaching}.
% [this is reflected in literature studying the increase in difficulty of learning with increase in reward sparsity (paper!)]
Plateaus during learning, which arise from saddle points in the loss landscape, have also been studied for (deep) neural networks in the supervised setting~\citep{saad1995line, dauphin2014identifying}, but do not arise in the supervised perceptron. Hence the RL setting can qualitatively change the learning trajectory. %The benefit of withholding penalties is that while slower, the perceptron reaches the highest level of expected reward in this case. This is a first example of a speed-accuracy trade-off that we will explore in more detail in \cref{sec:speed-acc} and that we also found in our experiments with Bossfight and Pong in \cref{sec:experiments}.

\textbf{Penalty:} The initial plateau can be reduced by providing a penalty or negative reward ($\eta_2>0$) when the student
fails in the task. This change provides weight updates much earlier in training and thus accelerates the escape from the plateau. The dynamics under this protocol are shown in~\cref{fig2}a. It is clear the penalty provides an initial speed-up in learning, as expected if the agent were to be unaligned and more likely to commit an error. However, a high penalty can create additional sub-optimal fixed points in the dynamics leading to a low asymptotic performance as seen in ~\cref{fig2}a (more on this in \cref{sec:phasediagram}). In the simulations, finite size effects occasionally permit escape from the sub-optimal fixed point and jumps to the optimal one, leading to a high variance in the results. The general form of the discounted reward $G_t$ in this case is given by eq. \ref{eq:discount}.

\textbf{Subtask:} The model is also able to capture the dynamics of more complicated protocols:~\cref{fig2}b shows learning under the protocol where a smaller sub-reward $r_b$ is received if the agent survives beyond a shorter duration $T_0 < T$ in addition to the final reward received for survival until timestep $T$, i.e. some reward is still received even if the agent does not survive for the entire episode. In this case $G_t = r_b\mathbb{I}(t \leq T_0)\prod_{t^{\prime}=1}^{T_0} \theta(y_{t^{\prime}}y^*_{t^{\prime}}) + \prod_{t^{\prime}=1}^T\theta(y_{t^{\prime}}y^*_{t^{\prime}})$ %We observe that receiving reward for surviving up to $T_0$ (in addition to the full survival reward) increases speed of convergence to the optimal accuracy with decreasing $T_0$. This is because reward is seen more frequently combined with the roll-back of reward (from the definition of $G_t$) to only `good' updates. With breadcrumb rewards, increasing $r_b$ increasing the initial speed of learning but at a cost of learning speed in the later stages of learning.

\textbf{Dense:} The model can capture scenarios in which rewards are densely received throughout an episode; this is reflected by the learning protocol where the agent receives a small reward $r_b$ for every correct decision made in an episode and a reward of 1 at time $T$ if the entire episode is successfully completed, i.e. like the previous method some reward is still received even if the agent does not survive for the entire episode, these dynamics are captured in~\cref{fig2}c. In this case the discounted reward $G_t = \sum_{t^{\prime}=1}^T r_b \theta(y_{t^{\prime}}y^*_{t^{\prime}}) + \prod_{t^{\prime}=t}^T\theta(y_{t^{\prime}}y^*_{t^{\prime}})$,

\subsection{Optimal hyper-parameter schedules}%
\label{sec:scheduling}

Hyper-parameter schedules are crucial for successful training of RL agents. In our setup, the two most important hyper-parameters are the learning rates and the episode length. In the RL perceptron, we can derive optimal schedules for both hyper-parameters. For simplicity, here we report the results in the spherical case, where the length of the student vector is fixed at $\sqrt D$  (we discuss the unconstrained case in the \cref{app:further_theory}), then $Q(\alpha)=1$ at all times and we only need to track the 
teacher-student overlap $\rho = R/\sqrt{Q}$, which quantifies the generalisation performance of the agent. Keeping the choice $\mathbb{I}(\Phi) = \prod_{t=1}^T\theta(y_ty_t^{*})$ we find the optimal schedules over episodes for $T$ and $\eta$ can then be found by maximising the change in overlap at each update, i.e. setting $\frac{\partial}{\partial T} \left(\frac{d\rho}{d\alpha}\right)$ and $\frac{\partial}{\partial \eta} \left(\frac{d\rho}{d\alpha}\right)$ to zero respectively. After some calculations, we find the optimal schedules to be
\begin{eqnarray}
        T_{\text{opt}} &&= \left\lfloor\frac{\sqrt{\pi}}{2}\frac{\eta \rho P}{(1-\rho^2)\sqrt{2Q}}\left[ 1 + \sqrt{1- \frac{\sqrt{2Q}}{\eta \rho} \frac{4(1-\rho^2)}{\sqrt{\pi}P\ln(P)} }\right ]\right\rfloor \nonumber \\
    &&\qq{and} \eta_{\text{opt}} = \sqrt{\frac{Q}{2\pi}}\frac{T(1-\rho^2)}{\rho P}
    \label{eq:schedules}
\end{eqnarray}
where $\lfloor\cdot\rfloor$ indicates the floor function. \Cref{fig3}a shows the evolution of $\rho$ under the optimal episode length schedule (dashed) compared to other constant episode lengths (green). Similarly, \cref{fig3}b shows the evolution of $\rho$ under the optimal learning rate schedule (dashed) compared to other constant learning rates (blue). The functional forms of $T_{\text{opt}}$ and $\eta_{\text{opt}}$ over time are shown in \cref{fig3}c.

Our analysis shows that a polynomial increase in the episode length gives the optimal performance in the RL perceptron, see \cref{fig3}c (top); increasing $T$ in the RL perceptron is akin to increasing task difficulty, and the polynomial scheduling of $T_{\text{opt}}$ specifies a curriculum. Curricula of increasing task difficulty are commonly used in RL to give convergence speed-ups and learn problems that otherwise would be too difficult to learn~\emph{ab initio}~\cite{narvekar2020curriculum}. 
Analogously, the fluctuations can be reduced by annealing the learning rate and averaging over a larger number of samples. Akin to work in RL literature studying adaptive step-sizes~\citep{Dabney2014ADAPTIVESF, NIPS2013_f64eac11}, we find that annealing the learning rate during training is beneficial for greater speed and generalisation performance. For the RL perceptron, a polynomial decay in the learning rate gives optimal performance as shown in \cref{fig3}c (bottom), consistent with work in the parallel area of high-dimensional non-convex optimization problems~\citep{dascoli2022optimal}, and stochastic approximation algorithms in RL~\citep{DBLP:journals/corr/DalalSTM17}.

\begin{figure*}
    \centering
    \includegraphics[width=0.9\textwidth]{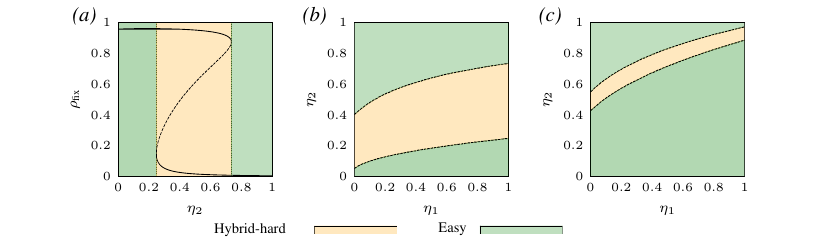}
    \caption{\textbf{Phase plots of learnability}. In the case where all decisions in an episode of length $T$ must be correct. \emph{(a)} the fixed points of $\rho$ for $T=13$ and $\eta_1=1$, the dashed portion of the line denotes where the fixed points are unstable. \emph{(b)} Phase plot showing regions of hardness for $T=13$. \emph{(c)} Phase plot showing regions of hardness for $T=8$. Green regions represent the \textit{Easy} phase where with probability 1 the algorithm naturally converges to the optimal $\rho_{\text{fix}}$ from random initialisation. The orange region indicates the \textit{Hybrid-hard} phase, where with high probability the algorithm converges to the sub-optimal $\rho_{\text{fix}}$ from random initialisation. \textit{Parameters: $D=900$, $Q=1$.}\vspace*{-\baselineskip}}
    \label{fig4}
\end{figure*}

\subsection{Phase space}%
\label{sec:phasediagram}
With a non-zero penalty ($\eta_2$), the generalisation performance of the agent can enter different regimes of learning. This is most clearly exemplified in the spherical case, where the number of fixed points of the ODE governing the dynamics of the overlap exist in distinct phases determined by the combination of reward and penalty. For the simplest case $\left(\mathbb{I}(\Phi) = \prod_t^T(y_ty^*_t)\right)$ these phases are shown in \cref{fig4}. \Cref{fig4}a shows the fixed points achievable over a range of penalties for a fixed $\eta_1 = 1$ (obtained from a numerical solution of the ODE in $\rho$). There are two distinct regions: 1) \textit{Easy}, where there is a unique fixed point and the algorithm naturally converges to this optimal $\rho_{\text{fix}}$ from a random initialisation, 2) a \textit{Hybrid-hard} region (given the analogy with results from inference problems \cite{ricci2019typology}), where there are two stable (1 good and 1 bad) fixed points, and 1 unstable fixed point, and either stable point is achievable depending on the initialisation of the student (orange). The `hybrid-hard' region separates two easy regions with very distinct performance levels. In this region the algorithm with high probability converges to $\rho_{\text{fix}}$ with the worse performance level. These two regions are visualised in ($\eta_1,\eta_2$) space in \cref{fig4}b for an episode length of $T=13$. The topology of these regions are also governed by episode length, with a sufficiently small T reducing the the area of the `hybrid-hard' phase to zero, meaning there is always 1 stable fixed point which may not necessarily give `good' generalisation. \Cref{fig4}c shows the phase plot for $T=8$, where the orange (hybrid-hard) has shrunk, this corresponds to the s-shaped curve in \cref{fig4}a becoming flatter (closer to monotonic). Details of the construction of \cref{fig4} are given in \cref{app:phase}. These regimes of learnability are not a peculiarity specific to the spherical case, indeed, we observe different regimes in the learning dynamics in the setting with unrestricted $Q$ which we report in \cref{app:further_theory}. We also show that using a logistic policy with the exact REINFORCE update indeed results in the RL-perceptron not being able to learn above some threshold of learning rates in \cref{app:reinforce_sims}a.

These phases show that at a fixed $\eta_1$ increasing $\eta_2$ will eventually lead to a first order phase transition, and the speed benefits gained from a non-zero $\eta_2$ will be nullified due to the transition into the hybrid-hard phase. In fact, when taking $\eta_2$ close to the transition point, instead of speeding up learning there is the presence of a critical slowing down, which we report in \cref{app:further_theory}. A common problem with REINFORCE is high variance gradient estimates leading to bad performance~\citep{Marbach2003ApproximateGM, https://doi.org/10.48550/arxiv.1506.02438}. The reward ($\eta_1$) and punishment ($\eta_2$) magnitude alters the variance of the updates, and we show that the interplay between reward, penalty and reward-condition and their effect on performance can be probed within our model. This framework opens the possibility for studying phase transitions between learning regimes \citep{Gamarnik_2022}.

\subsection{Speed-accuracy trade-off}%
\begin{figure}[b]
\includegraphics[width=0.3\textwidth]{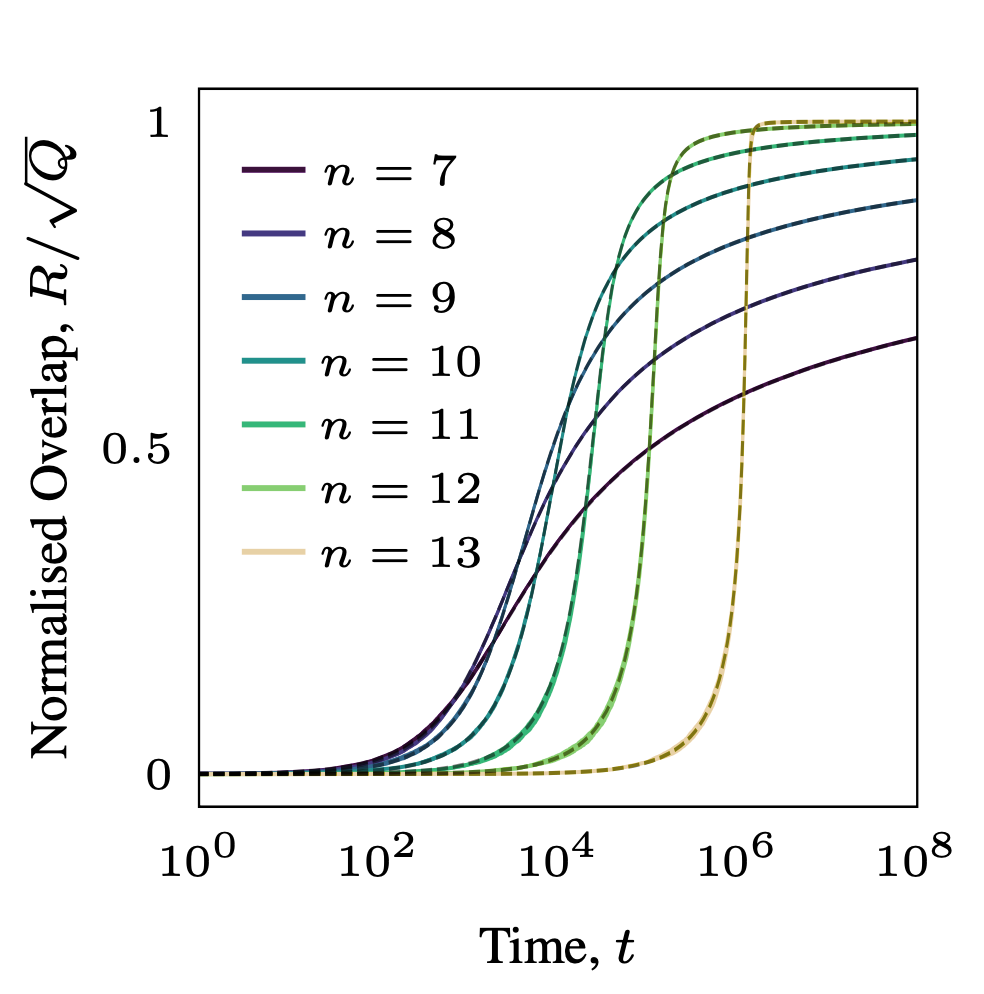}% Here is how to import EPS art
\caption{\label{fig5} \textbf{Speed accuracy tradeoff} Evolution of the normalised overlap between student and teacher for simulation (solid) and ODE solution (dashed) for the case where $n$ or more decisions in an episode of $T=13$ are required correct for an update with $\eta_2 = 0$. More stringent reward conditions slow learning
but can improve performance. \textit{Parameters}: $D = 900, \eta_1 = 1, \eta_2 = 0$.}
\end{figure}

\label{sec:speed-acc}
\Cref{fig5} shows the evolution of  normalised overlap $\rho = R/\sqrt{Q}$  between the student and teacher obtained from simulations and from solving the ODEs in the case where $n$ or more decisions must be correctly made in an episode of length $T=13$ in order to receive a reward (with $\eta_2=0$).
\begin{figure*}
    \centering
    \includegraphics[width=0.9\textwidth]{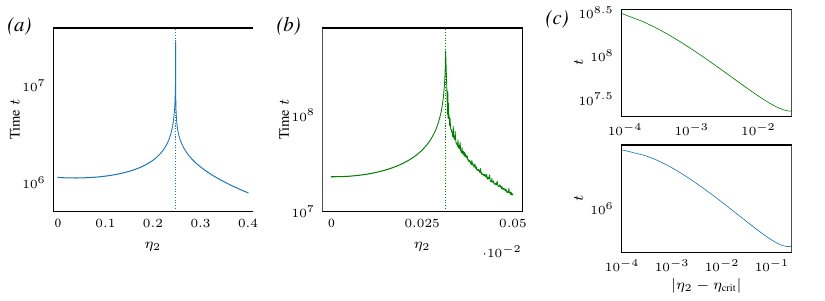}
    \caption{\textbf{Critical slowing down for $\eta_2 \neq 0$}. \emph{(a)} The time for convergence to the fixed point for $T=13$. \emph{(b)} The time for convergence to the fixed point for $T-=20$ \emph{(c)} Time for convergence plotted against distance of $\eta_2$ away from the critical penalty for $T=13$ (bottom) and $T=20$ (top). All plots are for the spherical case where the agent must get every decision correct in order to receive a reward of $\eta_1=1$, and receives a penalty of $\eta_2$ otherwise. \textit{Parameters: $D=900$, $Q=1$, $\eta_1=1$}\vspace*{-\baselineskip}}
    \label{suppfig3}
\end{figure*}
We observe a speed-accuracy trade-off, where decreasing $n$ increases the initial speed of learning but leads to worse asymptotic performance; this alleviates the initial plateau in learning seen previously in \cref{fig2}d at the cost of good generalisation. In essence, a lax reward function is probabilistically more achievable early in learning; but it rewards some fraction of incorrect decisions, leading to lower asymptotic accuracy. By contrast a stringent reward function slows learning but eventually produces a highly aligned student. For a given MDP, it is known that arbitrary shaping applied to the reward function will change the optimal policy (reduce asymptotic performance)~\citep{ng1999policy}. Empirically, reward shaping has been shown to speed up learning and help overcome difficult exploration problems~\citep{gullapalli1992shaping}. Reconciling these results with the phenomena observed in our setting is an interesting avenue for future work.

\subsection{Critical slowing down}

 With the addition of a penalty term we observe initial speed up in learning as shown in \cref{suppfig3}. Towards the end of learning, however, we observe a critical slowing down, and see how in many instances a non-zero $\eta_2$ can instead give an overall slowing to learning. This is most easily seen in the spherical case for the rule where all decisions in an episode of length $T$ must be correct for a reward: \cref{suppfig3}a shows the times to reach 0.99 of the fixed point starting from an initial $\rho=0$ for $T=13$ and $\eta_1=1$. We observe that increasing $\eta_2$ (up to $\eta_{\text{crit}}$, at which point the algorithm enters the hybrid-hard phase detailed in \cref{sec:phasediagram}) increases the time taken to reach the fixed point. This is similarly seen for $T=20$ in \cref{suppfig3}b. This slowing is not present over the entire range of $\eta_2$; it is true that for small values of $\eta_2$ there is actually a small speed up in the reaching of the fixed point, showing that the criticality severely reduces the range of $\eta_2$ that improves convergence speed. We plot the distance of of $\eta_2$ away from the critical penalty value ($|\eta_2-\eta_{\text{crit}}|$) against time for convergence in \cref{suppfig3}a, for $T=20$ (top) and $T=13$ (bottom). We observe a polynomial scaling of the convergence time with distance away from criticality.

\section{Experiments}%
\label{sec:experiments}

\begin{figure*}
    \centering
    \includegraphics[width=0.9\textwidth]{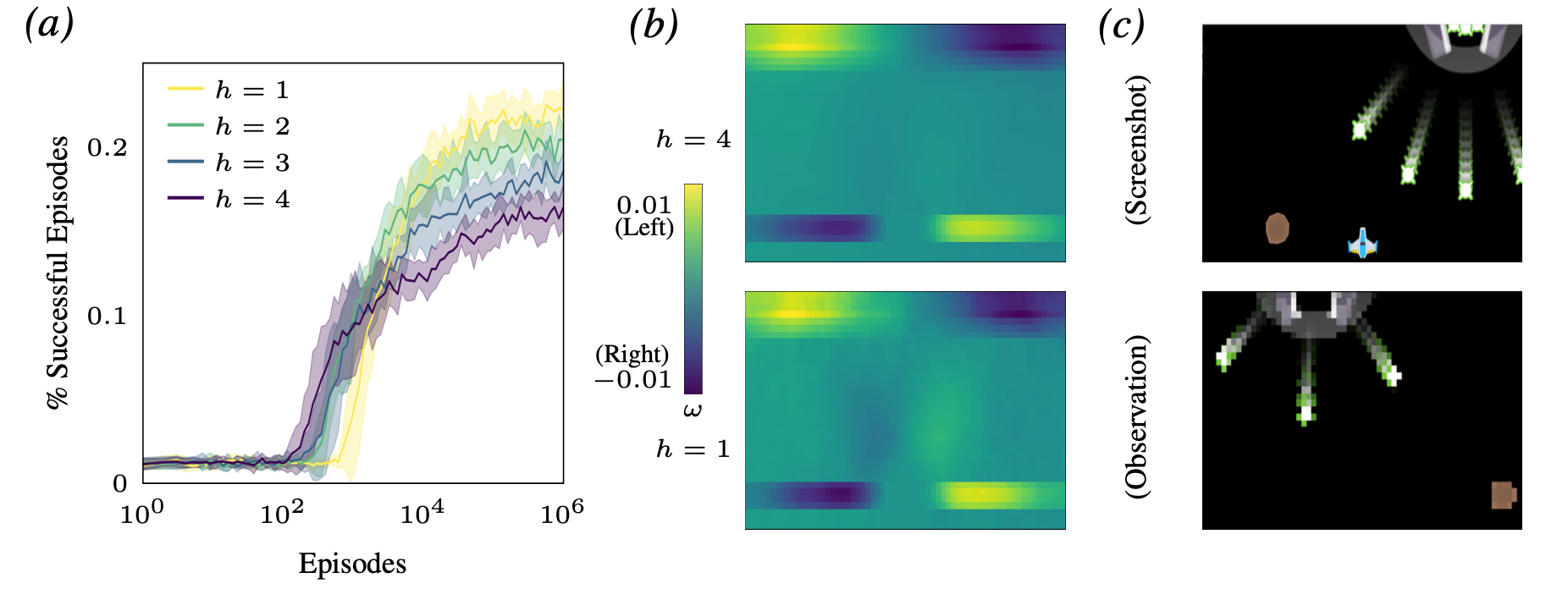}
    \caption{\rebuttal{\textbf{Empirical speed-accuracy tradeoff in Bossfight.} \emph{(a)} (Bossfight) Generalisation performance over training a perceptron policy network with REINFORCE algorithm, measured on evaluation episodes with $h=1$ lives. Agents trained in stringent conditions ($h=1$) learn slowly but eventually outperform agents trained in lax conditions ($h=4$), an instance of the speed-accuracy tradeoff. Shaded regions indicate SEM over 10 repetitions. \emph{(b)} (Bossfight) Policy network weights ($\omega$) for an agent with (top) $h=4$ lives and (bottom) $h=1$ life. For simplicity, one colour channel (red) is shown. Training with fewer lives increases the weight placed on dodging projectiles (see text). \emph{(c)} (Top) Example screenshot of a still of bossfight. (Bottom) Example observation that the policy network sees. \textit{Parameters: (Bossfight) -  $T=100,\eta_1=2e-3, \eta_2=0$.}}\vspace*{-\baselineskip}}
    \label{fig6}
\end{figure*}

\looseness=-1
To verify that our theoretical framework captures qualitative features of more general settings, we train agents from pixels on the Procgen~\citep{cobbe2019procgen} game `Bossfight'. To remain close to our theoretical setting, we consider a modified version of the game where the agent cannot defeat the enemy and wins only if it survives for a given duration $T$. On each timestep the agent has the binary choice of moving left/right and aims to dodge incoming projectiles. We give the agent $h$ lives, where the agent loses a life if struck by a projectile and continues an episode if it has lives remaining. This reward structure reflects the sparse reward setup from our theory and is analogous to requiring $n$ out of $T$ decisions to be correct within an episode. We further add asteroids at the left and right boundaries of the playing field which destroy the agent on contact, such that the agent cannot hide in the corners (see the screenshots of the game in~\cref{fig6}c). Observations, shown in~\cref{fig6}c (\textit{top}), are centred on the agent and downsampled to size $35\times 64 $ with three colour channels, yielding a $6720$ dimensional input. The pixels corresponding to the agent are set to zero since these otherwise act as near-constant bias inputs not present in our model~(\cref{fig6}c \textit{bottom}). The agent is endowed with a shallow policy network with logistic output unit that indicates the probability of left or right action. The weights of the policy network are trained using the exact REINFORCE policy gradient update (with additional entropy regularisation to steer away from an early deterministic policy).

To study the speed-accuracy trade-off, we train agents with different numbers of lives. As seen in \cref{fig6}a, we observe a clear speed-accuracy trade-off mediated by agent health, consistent with our theoretical findings (c.f. \cref{fig5}). \Cref{fig6}b shows the final policy weights for agents trained with $h=1$ and $h=4$. When compared to the game screenshots in~\cref{fig6}c, these show interpretable structure, roughly split into thirds vertically: the weights in the top third detect the position of the boss and centre the agent beneath it; this causes projectiles to arrive vertically rather than obliquely, making them easier to dodge. The weights in the middle third dodge projectiles. Finally, the weights in the bottom third avoid asteroids near the agent. Notably, the agent trained in the more stringent reward condition ($h=1$) places greater weight on dodging projectiles (seen from the stronger colours in \cref{fig6}b), showing the qualitative impact of reward on learned policy. Hence similar qualitative phenomena as in our theoretical model can arise in more general settings.

\begin{figure}[b]
\includegraphics[width=0.45\textwidth]{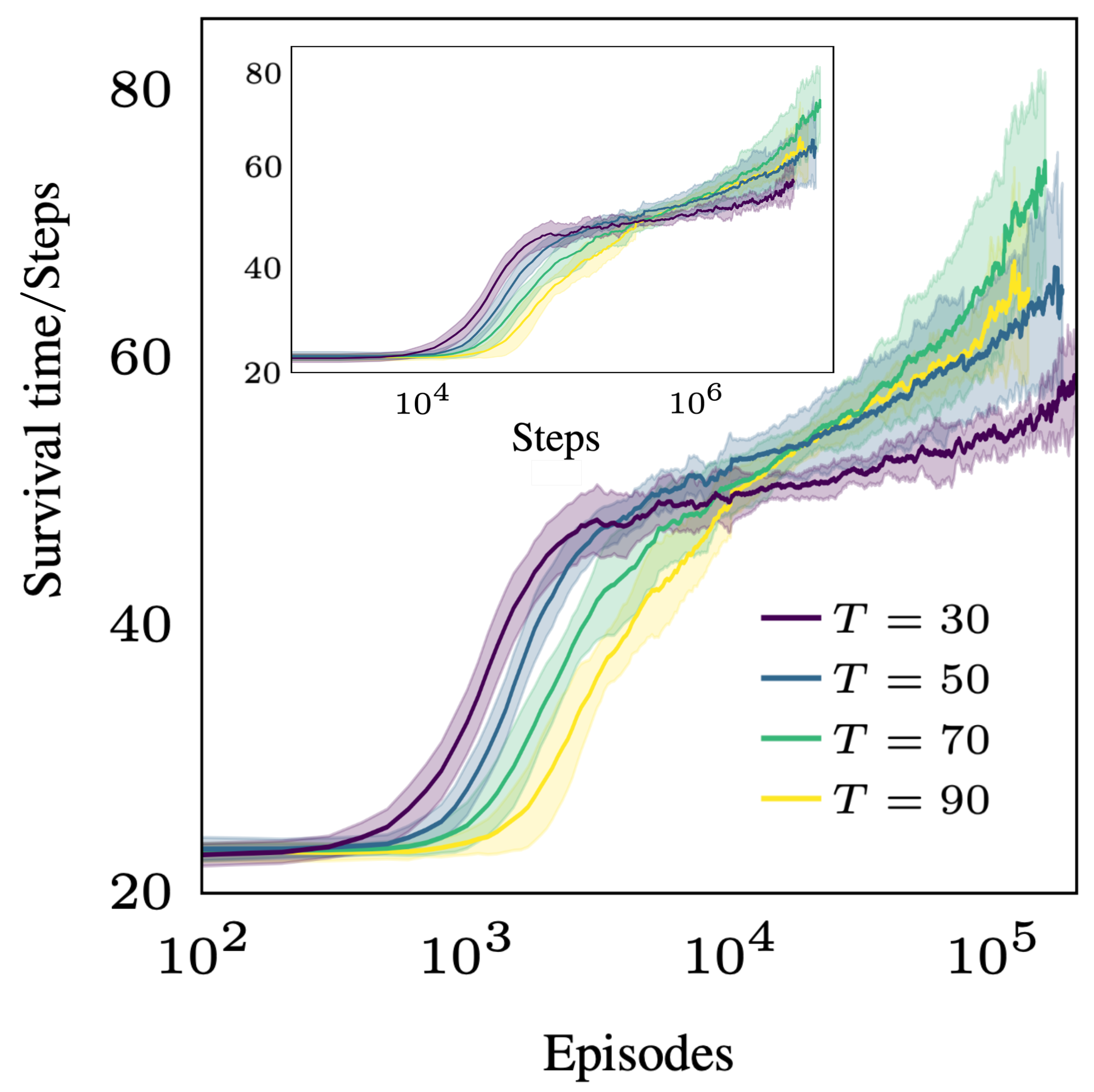}% Here is how to import EPS art
\caption{\rebuttal{ \textbf{Empirical speed-accuracy tradeoff in Pong.} The mean survival time over the course of training for agents required to survive up to completion of an episode of length $T$, in order to receive reward. Inset: Same data re-plotted for total number of training steps taken in the game. \textit{Parameters:} $\eta_1=2e-3, \eta_2 = 0$}}
\label{fig:8 sa_steps} 
\end{figure}
For a test of the generality of our conclusions, we train agents from pixels on the Arcade Learning Environment (ALE) \citep{machado18arcade} game `Pong' (using the exact REINFORCE update with a deep nonlinear network). The notion of lives (or requiring $n$ or more correct decisions in an episode for a reward) is essentially a way to control the difficulty of a task, whereby higher $n$ (fewer lives) is a more stringent condition i.e. a more difficult task. We examine a corresponding setup in Pong, where task difficulty is varied in order to study the dynamics of generalisation performance of agents. The Pong task difficulty is varied by changing the episode length $T$ which the agent must survive to in order to receive reward. Intuitively, larger $T$ is a more difficult task as an agent is required to survive for longer. On each timestep the agent has a binary choice of moving left/right and aims to return the ball. If the ball manages to get past the agent the episode ends without reward; if the agent survives until the end of the episode, it receives a reward. The decisions of the agent are sampled from the logistic output of a deep policy network, consisting of 2 convolutional layers, 2 fully connected layers and ReLU non-linearities with a sigmoidal output. Pong is deterministic, so in order to avoid memorisation, we introduce stochasticity by employing two approaches: 1. `Frame-skips'~\citep{kalyanakrishnan2021analysisframeskippingreinforcementlearning}, where whenever an action it is taken a random number of times; 2. Random initialisation, where a randomly selected pre-trained agent is run for a random number of timesteps in order to progress the game into a `random' initialisation state. The weights of the policy network are trained using the exact REINFORCE policy gradient update (with additional entropy regularisation) by running 20 agents in parallel. \rebuttal{\Cref{fig:8 sa_steps} shows a clear trend in speed of learning with more stringent reward conditions taking longer to learn. It is clear from \cref{fig:8 sa_steps} that asymptotic accuracy has not yet been reached within this time frame, unfortunately due to computational and time constraints it is not possible to train for longer. However, we can see the separation in the generalisation accuracy as agents trained to survive on episodes of $T=70$ begin to survive for longer at an increasing rate compared to agents trained on episodes of $T=50$, which in turn begin to survive for longer at an increasing rate compared to agents trained on episodes of $T=30$. Agents trained on episodes of $T=90$, however, are not seen to overtake in generalisation performance. Likely this is due to not training for long enough, it can be seen that agents trained on $T=90$ have not reached a regime where they are consistently receiving reward yet (and therefore updates are still infrequent). It would require more training steps for $T=90$ agents to more frequently receive reward and update weights, at which point we would expect an overtake in generalisation performance compared to the other agents.
The inset in \cref{fig:8 sa_steps} shows the same data as the main plot in \cref{fig:8 sa_steps}, but re-plotted against the number of total steps (steps of the agent in the game) instead of number of episodes. The trend remains, with an exaggerated speed difference but tighter difference in generalisation performance between the different agents (we give the results of statistical significance test that tests the significance of the mean performances being different from each-other in \cref{app:exper}). In practice, choosing whether to measure speed differences in number of episodes versus number of steps will be application dependent. For instance, suppose the bottleneck cost is taking a step, then step number is the appropriate measure. But if the bottleneck cost is the episode, e.g. due to having to return a robot to initial position, then episode number will be the appropriate measure. For more detailed experimental setup, see \cref{app:exper}.}

\section{Concluding perspectives}%
\label{sec:conclusion}
% \looseness=-1
%

The RL perceptron provides a framework to investigate high-dimensional policy gradient learning in RL for a range of plausible sparse reward structures. We derived closed ODEs that capture the \textit{average-case} learning dynamics in high-dimensional settings. The reduction of the high-dimensional learning dynamics to a low-dimensional set of differential equations permits a precise, quantitative analysis of learning behaviours: computing optimal hyper-parameter schedules, or tracing out phase diagrams of learnability.  Our framework offers a starting point to explore additional settings that are closer to many real-world RL scenarios, such as those with conditional next states. Furthermore, the RL perceptron offers a means to study common training practices, including curricula; and more advanced algorithms, like actor-critic methods. We hope to extract more analytical insights from the ODEs, particularly on how initialization and learning rate influence an agent's learning regime. Our findings emphasize the intricate interplay of task, reward, architecture, and algorithm in modern RL systems. 

We designed the RL perceptron to be the simplest possible model of reinforcement
learning that lends itself to an analytical treatment, so we limited the model
to binary action spaces, environmental states sampled from standard Gaussian
distributions, and simple shallow networks to learn the policy. We did not
consider state transitions that are conditioned on the action. While this simple
model already showed a rich behaviour, it is indeed possible to extend our model
to problems with more states, more realistic inputs, and more decisions using
universality results from statistical physics
\citep{mei2022generalization, goldt2020modeling, goldt2022gaussian,
  gerace2020generalisation, hu2022universality, dandi2023universality}. These
extensions would make it possible to define the notion of ``value'' on states/actions, meaning there is the potential to incorporate value-based RL algorithms or algorithms that combine policy and value-based methods, and this is a plan for future works. It would also be possible to extend to higher-dimensional action spaces instead of binary by considering work that finds learning curves for the multi-class perceptron \citep{Cornacchia_2023}. This again would widen the possibilities of RL agents we can consider, and also widen the number of suitable environments we can test against.

Policy learning is an key aspect of modern RL, but real-world applications require learning of both policies and value functions, which evaluate policies by assigning an expected reward to each (state, action) pair. Recently, \citet{bordelon2023loss} gave an analysis of the typical-case dynamics of the temporal difference algorithm to learn value functions using tools from statistical physics. The ultimate goal of a theory of reinforcement learning combining policy learning and the learning of value functions remains elusive due to the non-trivial interactions between the two processes, and presents an exciting challenge for further research.

\begin{acknowledgments}
We would like to thank Roberta Raileanu and Tim Rocktäschel for useful discussions.  This work was supported by a Sir Henry Dale Fellowship from the Wellcome Trust and Royal Society (216386/Z/19/Z) to A.S., and the Sainsbury Wellcome Centre Core Grant from Wellcome (219627/Z/19/Z) and the Gatsby Charitable Foundation (GAT3755). A.S. is a CIFAR Azrieli Global Scholar in the Learning in Machines \& Brains program. S.G. acknowledges co-funding from Next Generation EU, in the context of the National Recovery and Resilience Plan, Investment PE1 – Project FAIR “Future Artificial Intelligence Research, co-financed by the Next Generation EU [DM 1555 del 11.10.22].
\end{acknowledgments}
\appendix

\section*{Appendix}
\renewcommand{\thefigure}{A\arabic{figure}}
\section{POMDP form}%
\label{app_PG_update}
\setcounter{figure}{0}

The RL perceptron with its update rule \cref{eq:w_update} can be grounded in the (Partially Observable) Markov Decision Process (MDP) formalism where, at every timestep $t$, the agent occupies some state $s_t$ in the environment, and receives an observation $\mathbf{x}_t$ conditioned on $s_t$. An action $y_t$ is then taken by sampling from the policy $\pi(y_t \mid \mathbf{x}_t)$ (parameterised by the student $\mathbf{w}$), and the agent receives a reward accordingly. In the MDP framing, the observations $x_t \sim {N}(\mathbf{0}, \mathbf{\mathbbm{1}}_D)$ themselves are considered states. In the POMDP framing the observations are noisy representations of some low dimensional latent states: the state $s_t$ of the environment can be one of two states ${s_+, s_-}$ and $\mathbf{x}_t \sim P(\cdot|s_t)$ is a high dimensional sample representative of the underlying state, with $P(\cdot|s_\pm) = \mathcal{N}_\pm(\cdot | \mathbf{w^*})$. Where $\mathcal{N}_+(\cdot | \mathbf{w^*})$ is the ${N}(\mathbf{0}, \mathbf{\mathbbm{1}}_D)$ distribution, but
with zero-probability mass everywhere except in the half-space whose normal is parallel to $\mathbf{w^*}$, and $\mathcal{N}_-(\cdot | \mathbf{w^*})$ is  correspondingly non-zero in the half-space with a normal that is antiparallel to $\mathbf{w^*}$ --- (${N}(\mathbf{0}, \mathbf{\mathbbm{1}}_D)$ has been partitioned in two).
The next state $s_{t+1}$ is sampled with probability $P(s_{t+1}|s_t) \equiv P(s_{t+1}) = 1/2$ independently from the decision made by the student at previous steps. At the end of an episode, after all decisions have been made, we update the agent as in \cref{eq:w_update}. The POMDP and MDP framings are both equivalent, and lead to the same dynamics, as shown in \cref{app:derivations}. 

\section{Exact REINFORCE simulations}%
\label{app_exact_reinforce}
\renewcommand{\thefigure}{B\arabic{figure}}
\setcounter{figure}{0}
\begin{figure*}[!tp]
    \centering
\includegraphics[width=1.7in]{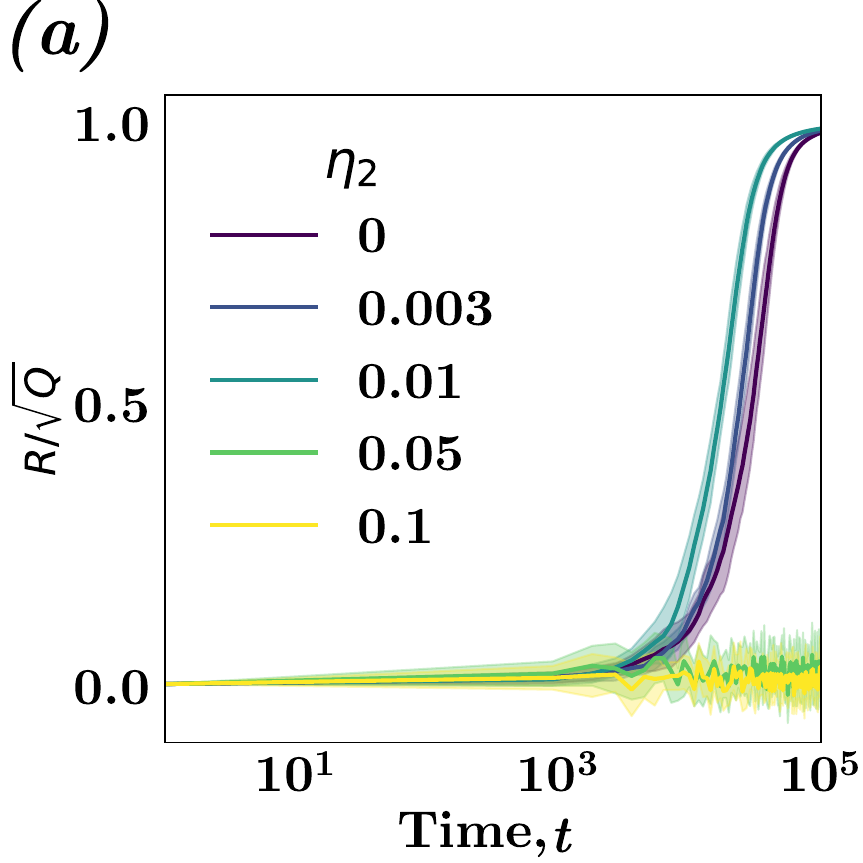}
\includegraphics[width=1.7in]{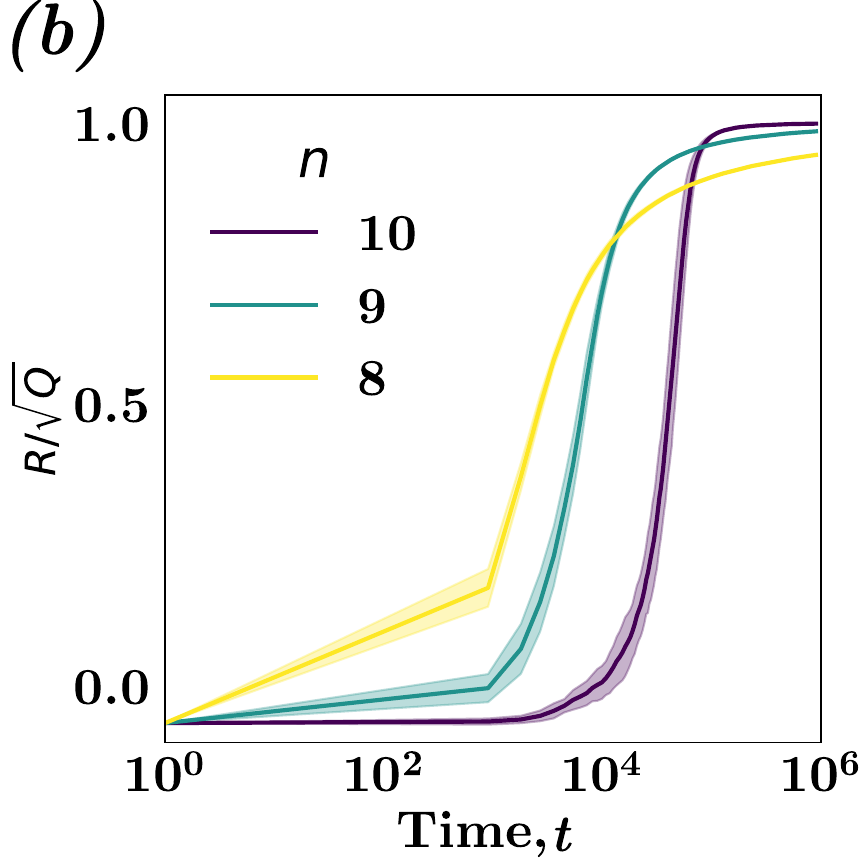}
\includegraphics[width=1.7in]{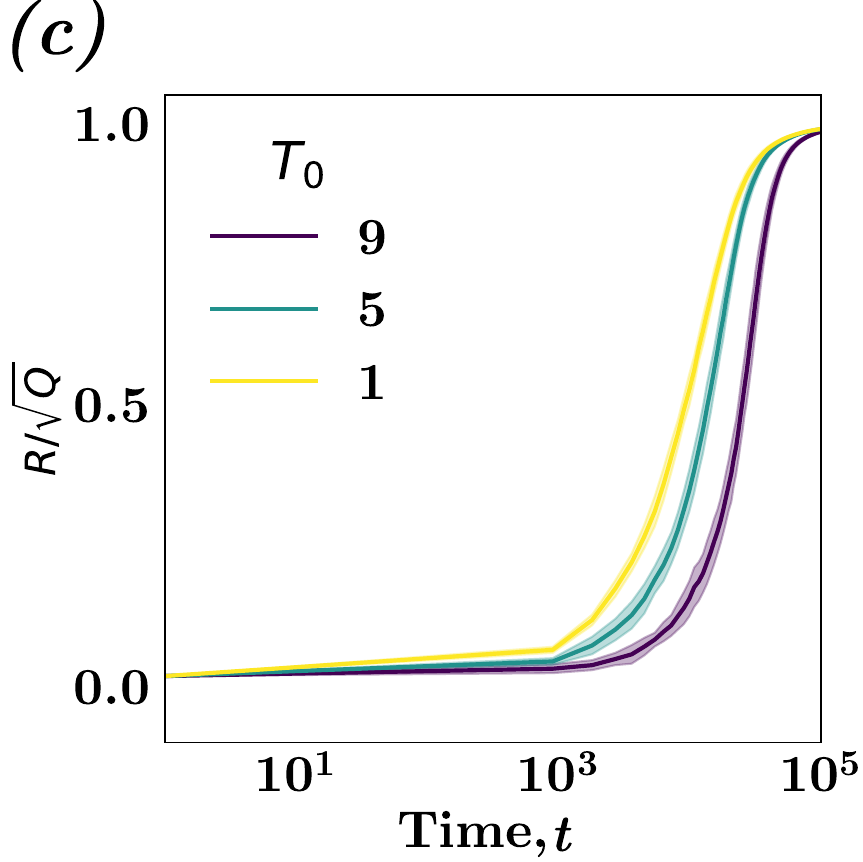}
\includegraphics[width=1.7in]{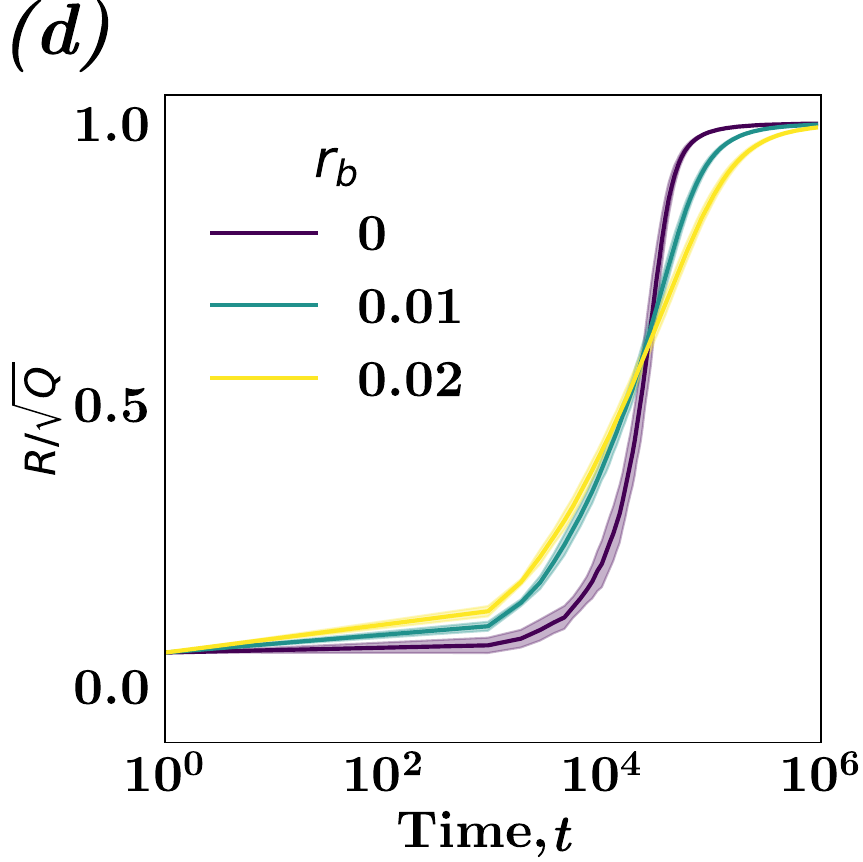}
\caption{\rebuttal{\textbf{Exact REINFORCE update is qualitatively consistent.} Evolution of the normalised student-teacher overlap $\rho$ using REINFORCE for simulation with actions sampled from a logistic policy with growth parameter $\Lambda$ in four reward protocols with episode length $T$ and reward $\eta_1$. \emph{(a)} All decisions in episode required correct for reward, with a negative reward $\eta_2$ otherwise. \emph{(b)} $n$ or more decisions required correct for reward. \emph{(c)} An additional reward of 0.2 is received at timestep $T_0$ if the agent survives beyond $T_0$ timesteps. \emph{(d)} An additional reward $r_b$ is received at timestep $t$ for every correct decision $y_t$ made in an episode. The corresponding figures to compare for the RL perceptron updates are: \emph{(a)}, \emph{(c)} and \emph{(d)} to \cref{fig2}a b and c respectively. And \emph{(b)} corresponds to \cref{fig5}.\textit{Parameters: $\Lambda=10000, T=10, \eta_1=1, D=900$}}}
\label{app:reinforce_sims}
\end{figure*}

\rebuttal{Simulations using the exact REINFORCE update were performed to show the validity of the RL-Perceptron update in capturing the qualitative behaviour of REINFORCE in our setting. We follow the standard RL-Perceptron setup detailed in \ref{setup_sec}, but in order to be able to take gradients, actions are sampled instead from a logistic policy $\pi_\bw(a_t |\bx_t) = \left(1 + \text{exp}\left(-\Lambda a_t \frac{\mathbf{w}^\intercal \mathbf{x}_t}{\sqrt{D}} \right) \right)^{-1}$ for large $\Lambda$. In the large $\Lambda$ limit the deterministic policy $\pi_\bw(a_t |\bx_t) = \frac{1}{2}(1 + a_t\sgn(\frac{\bw^\intercal \bx_t}{\sqrt{D}}))$ is well approximated. And instead of the RL-perceptron update \cref{eq:w_update}, we compute the true policy gradient and perform the REINFORCE update. \Cref{app:reinforce_sims} shows the evolution of the normalised overlap between the teacher and student for 4 various learning scenarios. \Cref{app:reinforce_sims}a shows the case where a negative reward is received if not all decisions in an episode are correctly made. It is consistent with \cref{fig2}a where an increase of the size of negative reward gives an initial speed-up in learning before learning fails altogether (which is also consistent with the results of \cref{sec:phasediagram}) and \cref{app:further_theory}, where above some threshold $\eta_2$ learning fails). \Cref{app:reinforce_sims}b shows the reward scheme where $n$ or more decisions in an episode are required correct for a reward; there is a speed accuracy tradeoff consistent with \cref{fig5}. \Cref{app:reinforce_sims}c and \cref{app:reinforce_sims}d show the subtask and dense reward protocols (\cref{setup_sec}) respectively. The shapes and ordering of the curves show good agreement with \cref{fig2}b and \cref{fig2}c.}

\section{Derivations}
\label{app:derivations}
\renewcommand{\thefigure}{C\arabic{figure}}
\setcounter{figure}{0}

\textbf{Thermodynamic Limit}: In going from the stochastic evolution of the state vector $\mathbf{w}$ to the deterministic dynamics of the order parameters, we must take the thermodynamic limit. For the ODE involving $R$ we must take the inner product of \cref{eq:w_update} with $\mathbf{w^*}$.

\begin{eqnarray}
    \mathbf{w}^{(\mu+1)} &&= \mathbf{w}^{(\mu)}+\frac{\eta}{T\sqrt{D}} \sum_{t=1}^{T}y_t^{(\mu)}  \mathbf{x}_{t}^{(\mu)}  G_t^{(\mu)} \\
    DR^{(\mu+1)}  &&= DR^{(\mu)}+\frac{\eta}{T\sqrt{D}}\sum_{t=1}^{T}y_t^{(\mu)}  \mathbf{w^*}^\intercal\mathbf{x}_{t}^{(\mu)}  G_t^{(\mu)}
    \label{eq:pre_R}
\end{eqnarray}

From this point, for ease of notation, any variable(s) with a subscript $\mu$ refers to said variable(s) in the $\mu$th episode. We subtract $DR^\mu$ from \ref{eq:pre_R} and sum over $l$ episodes, the LHS is a telescopic sum, and \ref{eq:pre_R} becomes

\begin{eqnarray}
     \frac{D(R^{\mu+l}-R^\mu)}{l}  &&= \frac{\eta}{T\sqrt{D}}\frac{1}{l}\sum_{i=0}^{l-1}\left(\sum_{t=1}^{T}y_t  \mathbf{w^*}^\intercal\mathbf{x}_{t}  G_t\right)^{\mu+i} \label{discrete_r} \\
     \frac{dR}{d\alpha} &&= \frac{\eta}{T}\left\langle \sum_{t=1}^{T}y_t  \frac{\mathbf{w^*}^\intercal\mathbf{x}_{t}}{\sqrt{D}} G_t\right\rangle \label{dr} \\
     &&= \frac{\eta}{T}\left\langle \sum_{t=1}^{T} \text{sgn}(\lambda_t) \nu_t  G_t\right\rangle
     \label{finalr}
\end{eqnarray}

We go from eq. \ref{discrete_r} to eq. \ref{dr} by taking the limit $D \rightarrow \infty$, $l \rightarrow \infty$ and $l/D = d\alpha \rightarrow 0$. The RHS of eq. \ref{discrete_r} is a sum of a large number of random variables, and by the central limit theorem is self-averaging in the thermodynamic limit (under the assumption of weak correlations between episodes), consequently the LHS is self-averaging. And we go from \ref{dr} to \ref{finalr} by considering the \textit{aligning fields} defined in~\cref{sec:ode}.  A similar procedure can be followed for order parameter $Q$, but we instead take the square of \cref{eq:w_update} and go to the limit described, obtaining:

\begin{eqnarray}
    DQ^{\mu+1} &&= DQ^\mu + \frac{2\eta}{\sqrt{D}}\left(\frac{1}{T} \sum_{t=1}^{T}y_t  \mathbf{w}^\intercal\mathbf{x}_{t}  G_t\right)^\mu \nonumber \\ 
    &&+ \frac{\eta^2}{ D}\left( \frac{1}{T^2}\sum_{t, t^{\prime}=1}^{T} y_{t} y_{t^{\prime}}\mathbf{x}_{t}^\intercal \mathbf{x}_{t^{\prime}} G_t G_{t^\prime} \right)^\mu \nonumber \\
    \frac{D(Q^{\mu+l}-Q^\mu)}{l} &&= \frac{2\eta}{T}\frac{1}{l}\sum_{i=0}^{l-1}\left(\sum_{t=1}^{T}y_t \frac{\mathbf{w}^\intercal\mathbf{x}_{t}}{\sqrt{D}}  G_t\right)^{\mu+i} \nonumber \\
    &&+ \frac{\eta^2}{T^2} \frac{1}{l}\sum_{i=0}^{l-1} \left( \sum_{t, t^{\prime}=1}^{T} y_{t} y_{t^{\prime}}\frac{\mathbf{x}_{t}^\intercal \mathbf{x}_{t^{\prime}}}{D} G_t G_{t^\prime} \right)^{\mu+i} \\
    \frac{dQ}{d\alpha} &&= \frac{2\eta}{T}\left\langle \sum_{t=1}^{T} \text{sgn}(\lambda_t) \lambda_t  G_t\right\rangle \nonumber \\
    &&+ \frac{\eta^2}{T^2} \left\langle \sum_{t=1}^{T} G_t^2\right\rangle + \mathcal{O}\left( \frac{1}{D}\right)
    \label{dq}
\end{eqnarray}

We have now obtained a general set of ODEs describing the learning dynamics. $G_t$ is general and will depend on the environment specific condition for reward. For the case of sparse reward received at the end of episode ($ G_t = r_1\mathbb{I}(\Phi(y_{1:T}, y^*_{1:T})) -r_2 (1-\mathbb{I}(\Phi(y_{1:T}, y^*_{1:T})))$). And eqs. \ref{finalr} and \ref{dq} become

\begin{eqnarray}
     \frac{dR}{d\alpha} &&= \frac{\eta_{1}+\eta_2}{T}\left\langle \sum_{t=1}^{T}\nu_t \mathrm{sgn}(\lambda_t)  \mathbb{I}(\Phi)\right\rangle-\eta_{2}\left\langle\nu \mathrm{sgn}(\lambda) \right\rangle \label{truedr}\\
    \frac{dQ}{d\alpha} &&=\frac{2\left(\eta_{1}+\eta_{2}\right)}{T }\left\langle\sum_{t=1}^{T} \lambda_t \mathrm{sgn}(\lambda_t)  \mathbb{I}(\Phi)\right\rangle-2 \eta_{2}\left\langle \lambda \mathrm{sgn}(\lambda) \right\rangle \nonumber \\
    &&+\frac{\eta_{1}^{2}-\eta_{2}^{2}}{T^{2} D}\left\langle\sum_{t, t^{\prime}=1}^{T} y_{t} y_{t^{\prime}}\mathbf{x}_{t}^\intercal \mathbf{x}_{t^{\prime}}  \mathbb{I}(\Phi)\right\rangle \nonumber \\
    &&+\frac{\eta_{2}^{2}}{T^{2} D}\left\langle\sum_{t, t^{\prime}=1}^{T} y_{t} y_{t^{\prime}}\mathbf{x}_{t}^\intercal \mathbf{x}_{t^{\prime}} \right\rangle. \label{truedq}
\end{eqnarray}

\textbf{Computing Averages}: It remains to compute the expectations. Recalling the definitions:

\begin{align}
    \nu = \frac{\mathbf{w^*}^\intercal \mathbf{x}}{\sqrt{D}} \quad \mathrm{ and } \quad \lambda = \frac{\mathbf{w}^\intercal \mathbf{x}}{\sqrt{D}},
    \label{eq:fields}
\end{align}

which are sums of $N$ independent terms and by the central limit theorem they obey a Gaussian distribution, we note
\begin{align}
    \langle \nu \rangle &= \langle \lambda \rangle = 0 \\
    \langle \nu ^ 2 \rangle &= 1 \quad \mathrm{, } \quad \langle \lambda^2 \rangle = Q\\
     \langle \nu \lambda \rangle &= \frac{\mathbf{w^*}^\intercal\mathbf{w}}{D} = R
\end{align}

All expectations in the ODEs can be expressed in terms of the constituent expectations given below (which are trivially computed by considering the Gaussianity of the above variables):

\begin{eqnarray}
    \left \langle \nu \mathrm{sgn}(\lambda)  \right \rangle = \sqrt{\frac{2}{\pi}}\frac{R}{\sqrt{Q}} \mathrm{, }\quad \left \langle \lambda \mathrm{sgn}(\lambda) \right \rangle = \sqrt{\frac{2Q}{\pi}} \mathrm{, } \nonumber \\
    \left \langle \nu \mathrm{sgn}(\nu)  \right \rangle = \sqrt{\frac{2}{\pi}} \mathrm{, }\quad \left \langle \lambda \mathrm{sgn}(\nu) \right \rangle = \sqrt{\frac{2}{\pi}}R
    \label{eq:mfs}
\end{eqnarray}

\begin{align}
    &\frac{1}{D}\left\langle\sum_{t, t^{\prime}=1}^{T}  y_{t} y_{t^{\prime}}\mathbf{x}^\intercal_{t} \mathbf{x}_{t^{\prime}}\right\rangle \nonumber \\
    &= \frac{1}{D}\left\langle\left(\sum_{t=1}^{T} \mathbf{x}^\intercal_{t} \mathbf{x}_{t}+2 \sum_{t=2}^{T} \sum_{t^{\prime}=1}^{t-1}y_{t} y_{t^{\prime}}\mathbf{x}^\intercal_{t} \mathbf{x}_{t^{\prime}}\right)\right\rangle \nonumber\\
    &= T + \mathcal{O}(1/D) \label{cross}
\end{align}

The terms involving $\Phi$ will in general consist of expectations containing step functions $\theta(x)$ (1 for $x > 0$, 0 otherwise), specifically $\theta(\nu \lambda)$ (1 if student decision agrees with teacher, 0 otherwise) and $\theta(-\nu \lambda)$ (1 if student decision disagrees with teacher, 0 otherwise). When we encounter these terms, they can be greatly simplified by considering the following equivalences:
\begin{eqnarray}
    \mathrm{sgn}(\lambda)\theta(\nu \lambda) = \frac{1}{2}(\mathrm{sgn}(\lambda) + \mathrm{sgn}(\nu)) \\
    \mathrm{ and } \quad  \mathrm{sgn}(\lambda)\theta(-\nu \lambda) = \frac{1}{2}(\mathrm{sgn}(\lambda) - \mathrm{sgn}(\nu)) 
\end{eqnarray}

We show as an example the case where $\Phi$ is the condition to get all decisions correct in an episode, $\mathbb{I}(\Phi) = \prod_{t=1}^T\theta(\nu_t \lambda_t)$, where $\theta(x)$ is the step function (1 for $x > 0$, 0 otherwise). The first term in Eq. \ref{truedr} can be addressed:

\begin{align}
    \left\langle\frac{1}{T} \sum_{t=1}^{T}\nu_t \mathrm{sgn}(\lambda_t) \mathbb{I}(\Phi)\right\rangle &\rightarrow \left\langle\frac{1}{T} \sum_{t=1}^{T} \nu_t \mathrm{sgn}(\lambda_t)   \prod_{s=1}^T\theta(\nu_s \lambda_s)\right\rangle \\
    &= \left\langle \nu_t \mathrm{sgn}(\lambda_t) \theta(\nu_t \lambda_t)\right \rangle \left \langle \prod_{s\neq t}^T\theta(\nu_s \lambda_s) \right\rangle \\
    &= \frac{1}{2}\left \langle \nu_t (\mathrm{sgn}(\lambda_t) + \mathrm{sgn}(\nu_t)) \right \rangle P^{T-1}\label{mix_1}\\
    &= \frac{1}{\sqrt{2\pi}} \left (1 + \frac{R}{\sqrt{Q}} \right)P^{T-1} 
\end{align}

where $P$ is the probability of making a single correct decision, and can be calculated by considering that an incorrect decision is made if $\mathbf{x}$ lies in the hypersectors defined by the intersection of $\mathcal{N}_\pm(\cdot | \mathbf{w^*})$ and $\mathcal{N}_\pm(\cdot | \mathbf{w})$, the angle $\epsilon$ subtended by these hypersectors is equal to the angle between $\mathbf{w^*}$ and $\mathbf{w}$.

\begin{equation}
    P = \left( 1- \frac{\epsilon}{\pi} \right) = \left( 1- \frac{1}{\pi}\text{cos}^{-1}\left(\frac{R}{\sqrt{Q}} \right) \right)
\end{equation}

Similarly, the first term in Eq. \ref{truedq} can be addressed:

\begin{align}
    &\left\langle\frac{2}{T} \sum_{t=1}^{T} \lambda_t \mathrm{sgn}(\lambda_t)  \prod_{s=1}^T\theta(\nu_s \lambda_s)\right\rangle \nonumber \\
    &= \left \langle \lambda_t (\mathrm{sgn}(\lambda_t) + \mathrm{sgn}(\nu_t)) \right \rangle P^{T-1} \nonumber \\
    &= \sqrt{\frac{2Q}{\pi}} \left (1 + \frac{R}{\sqrt{Q}} \right)P^{T-1} \label{mix_2}
\end{align}

The cross terms in \ref{truedq} can also be computed: 

\begin{eqnarray}
    &&\frac{1}{D}\left\langle\sum_{t, t^{\prime}=1}^{T} y_{t} y_{t^{\prime}}\mathbf{x}_{t}^\intercal \mathbf{x}_{t^{\prime}}  \prod_{s=1}^T\theta(\nu_s \lambda_s)\right\rangle \nonumber \\
    &&= \frac{1}{D}\left\langle\left(\sum_{t=1}^{T} \mathbf{x}^\intercal_{t} \mathbf{x}_{t}+2 \sum_{t=2}^{T} \sum_{t^{\prime}=1}^{t-1}y_{t} y_{t^{\prime}}\mathbf{x}^\intercal_{t} \mathbf{x}_{t^{\prime}}\right)\prod_{s=1}^T\theta(\nu_s \lambda_s)\right\rangle \nonumber \\
    &&= TP^T + \mathcal{O}(1/D)
\end{eqnarray}

where the 2nd term can be neglected in the high dimensional limit. Substituting these computed averages into equations \ref{truedr} and \ref{truedq}, the ODEs for the order parameters can be written:

\begin{eqnarray}
     \frac{dR}{d\alpha} &&= \frac{\eta_1 + \eta_2}{\sqrt{2\pi}}\left (1 + \frac{R}{\sqrt{Q}} \right)P^{T-1} - \eta_2 R \sqrt{\frac{2}{\pi Q}} \\
     \frac{dQ}{d\alpha} &&= (\eta_1 + \eta_2)\sqrt{\frac{2Q}{\pi}}\left (1 + \frac{R}{\sqrt{Q}} \right) P^{T-1} \nonumber \\
     &&- 2\eta_2\sqrt{\frac{2Q}{\pi}} +  \frac{(\eta_1^2 - \eta_2^2)}{T}P^T + \frac{\eta_2^2}{T}
\end{eqnarray}

\textbf{Equivalence of POMDP formulation}

The ODEs governing the dynamics of the order parameters in the previous section can be equivalently calculated under the formulation involving the underlying states $\{s_+, s_-\}$ defined in section ~\ref{setup_sec}. The underlying system can take a multitude of trajectories ($\tau$) in state space, there are $2^T$ trajectories in total (as the system can be in 2 possible states at each timestep), and expectations must now include the averaging over all possible trajectories. All expectations will now be of the following form, where the dot ($\cdot$) denotes some arbitrary term to be averaged over.

\begin{align}
    \left \langle \cdot \right \rangle = \sum_{\tau} \mathrm{P}(\tau) \left \langle \cdot \mid \tau \right \rangle
\end{align}

By considering symmetry of the Gaussian and `half-Gaussian' ($\mathcal{N}_\pm$) distributions, all expectations in \ref{eq:mfs} can be seen to be identical regardless of whether expectations are taken with respect to the full Gaussian or the half-Gaussian distributions, i.e.

\begin{align}
    \langle \cdot \rangle_{\mathcal{N}} = \langle \cdot \rangle_{\mathcal{N}_+} =\langle \cdot \rangle_{\mathcal{N}_-} 
\end{align}

this implies that all expectations are independent of the trajectory of the underlying system, hence averaging over all trajectories leaves all expectations unchanged. This also allows the extension to arbitrary transition probabilities between the underlying states $\{s_+, s_-\}$. 

\textbf{Other Reward structures}

The expectations can be calculated in other conditions of $\Phi$ from considering combinatorical arguments. We state the ODEs for two reward conditions.

\textbf{$\mathbf{n}$ or more}: the case where $\Phi$ is the requirement of getting $n$ or more decisions in an episode of length $T$ correct. We give the ODEs below for the case of $\eta_2 = 0$

\begin{eqnarray}
    \frac{dR}{d\alpha} &&= \frac{\eta_1 }{T\sqrt{2\pi}} \sum_{i=n}^T \binom{T}{i} \biggl[ i \left (1 + \frac{R}{\sqrt{Q}} \right)(1-P) \nonumber \\
    &&- (T-i)\left(1 - \frac{R}{\sqrt{Q}}  \right)P \biggr] P^{i-1} \left( 1-P\right )^{T-i-1}\\
     \frac{dQ}{d\alpha} &&= \frac{\eta_1 }{T}\sqrt{\frac{2Q}{\pi}} \sum_{i=n}^T \binom{T}{i} \biggl[ i\left (1 + \frac{R}{\sqrt{Q}} \right)(1-P) \nonumber \\
     &&- (T-i)\left(1 - \frac{R}{\sqrt{Q}}  \right)P\biggr]P^{i-1} \left( 1-P\right )^{T-i-1} \nonumber \\
     &&+ \frac{\eta_1^2}{T} \sum_{i=n}^T \binom{T}{i} P^{i} \left(1-P\right )^{T-i}
\end{eqnarray}

\textbf{Breadcrumb Trails} We also consider the case where a reward of size $\eta_1$ is received if all decisions in an episode are correct in addition to a smaller reward of size $\beta$ for each individual decision correctly made in an episode:

\begin{eqnarray}
    \frac{dR}{d\alpha} &&=  \frac{1}{\sqrt{2\pi}}\left[\left (1 + \frac{R}{\sqrt{Q}} \right)\left(\eta_1 P^{T-1} + \beta \right) + \beta (T-1)\frac{R}{\sqrt{Q}}P\right] \nonumber \\
    \frac{dQ}{d\alpha} &&= \sqrt{\frac{2Q}{\pi}}\left[\left (1 + \frac{R}{\sqrt{Q}} \right)\left(\eta_1 P^{T-1} + \beta \right) + 2\beta (T-1)P\right] \nonumber \\
    &&+ \left(\eta_1^2 + (T+1)\eta_1\beta\right)\frac{P^{T}}{T} \nonumber \\
    &&+ \beta^2(T+1)\left ( \frac{1}{2} + \frac{1}{3}(T-1)P\right)\frac{P}{T}
\end{eqnarray}

\section{Unconstrained Simulations}\label{app:further_theory}
\renewcommand{\thefigure}{D\arabic{figure}}
\setcounter{figure}{0}

\begin{figure}[b]
    \includegraphics[width=0.48\textwidth]{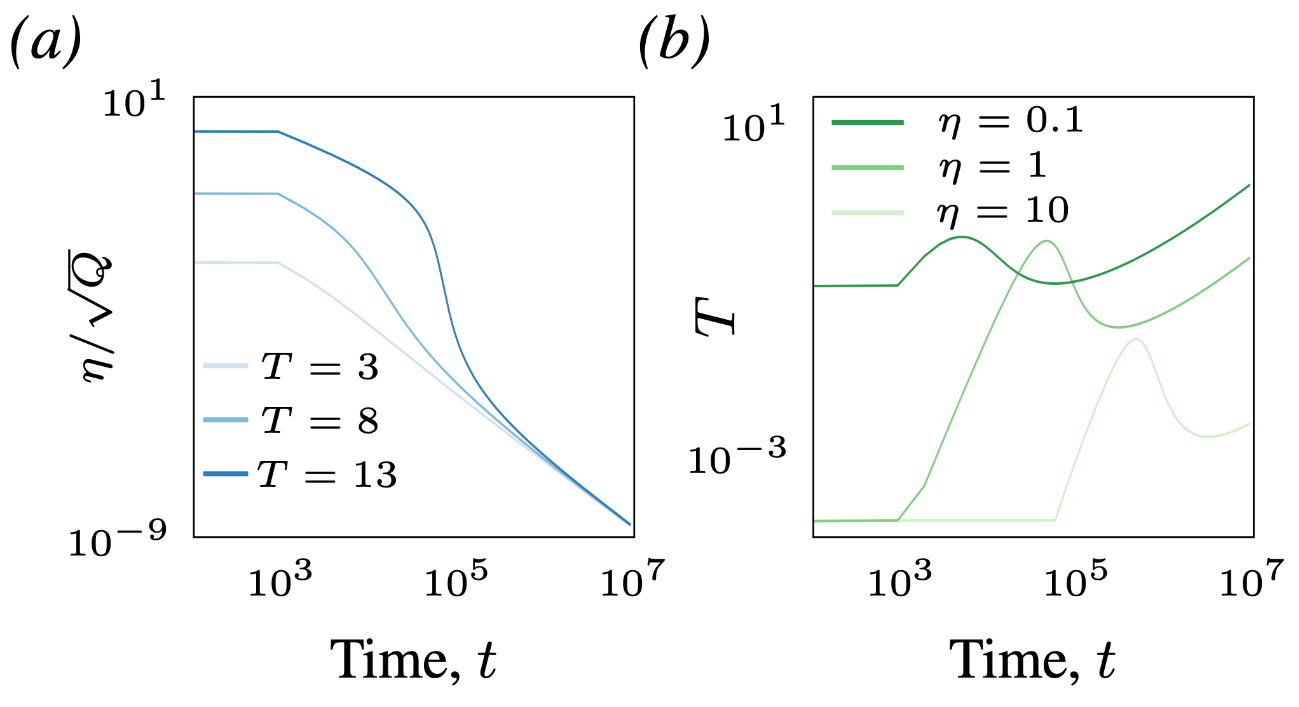}
    \caption{\textbf{Optimal Schedules for the unconstrained student}. \emph{(a)} Evolution of optimal $\eta$ \emph{(a)} and $T$ \emph{(b)} over learning, while following the specified optimal schedule, over a range of rewards and episode lengths. \textit{Parameters: $D=900$, $\eta_2=0$, (a) $T=8$, (b) $\eta=1$.}}
    \label{suppfig1}
\end{figure}
\textbf{Optimal scheduling:} The optimal schedules for learning rate and episode length (\cref{eq:schedules}) hold in the unconstrained case
too (where $Q(\alpha)$ isn't restricted to the surface of a sphere); this is 
because the parameters were derived from the general requirement of 
extremising the update of $\rho$ from any point in the $(\rho, Q)$ plane. 
The evolution of $T_{\text{opt}}$ and $\eta_{\text{opt}}$ over time (while following their respective scheduling) is shown in \cref{suppfig1}. In the unconstrained case the magnitude of the student grows quadratically, an increase $Q$ acts as a decrease in effective learning rate. Hence contrary to the spherical case a decaying learning rate is not optimal, and optimal $T$ grows much slower, as shown in \cref{suppfig1}b; the plots for $T_{\text{opt}}$ do not show a clear trend and require further investigation. The evolution of $\eta_{\text{opt}}/\sqrt{Q}$ is plotted in \cref{suppfig1}a, this value is the effective learning rate and we observe a polynomial decay in the value as with the spherical case presented in \cref{sec:scheduling}.

\begin{figure*}
    \centering
    \includegraphics[width=\textwidth]{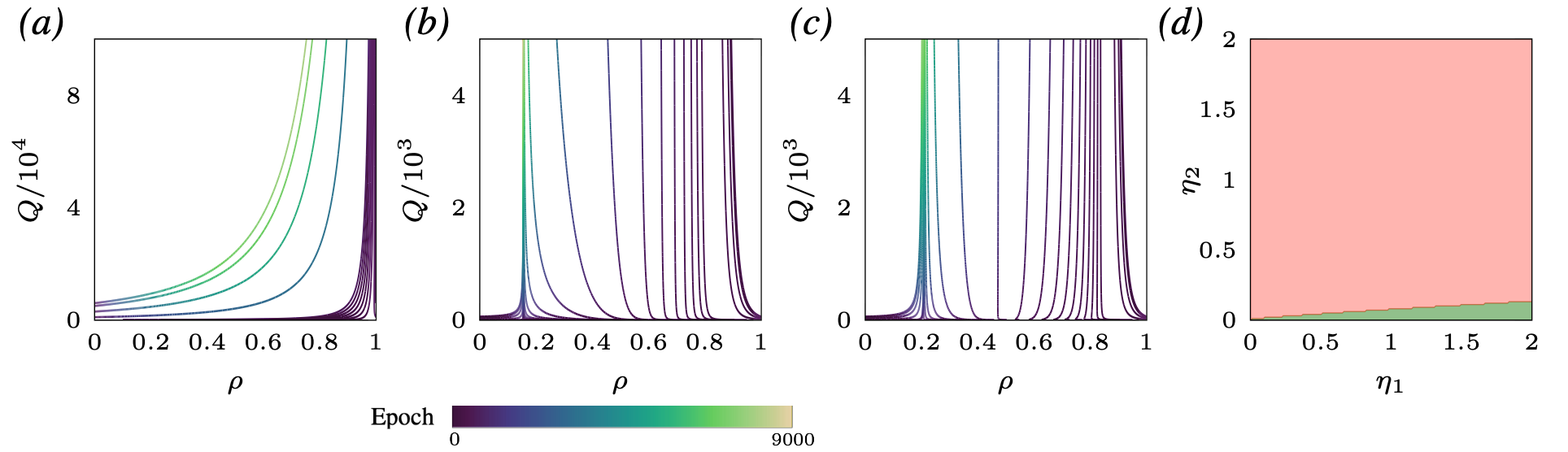}
    \caption{\textbf{Unconstrained flow and phase plots for increasing size of negative reward}.  Flow in the $(\rho,Q)$ plane (flow goes in the direction of increasing $Q$) for the case where all decisions in an episode are required correct for an reward of $\eta_1=1$ and a penalty other wise of $\eta_2 = $ 0 \emph{(a)}, 0.05  \emph{(b)}, and 0.045 \emph{(c)}. \emph{(d)} Phase plot showing the region where learning failed (red) and succeeded (green) over the ($\eta_1, \eta_2$) plane, for the same learning rule. \textit{Parameters: Initialised from $\rho=0$ and $Q=1$}}
    \label{suppfig2}
\end{figure*}

\textbf{Phases:} The phases observed in \cref{fig4} are not an artifact of the spherical case. When $Q(\alpha)$ is not constrained we also observe regimes where a `bad' fixed point of $\rho$ may be attained. \Cref{suppfig2} shows flow diagrams in the $(\rho, Q)$ plane for various parameter instantiations in the case where a reward of $\eta_1 = 1$ is received if all decisions in an episode of length $T=8$ are correctly made, and a penalty of $\eta_2$ otherwise. \Cref{suppfig2}a is the flow diagram for $\eta_2=0$, in this regime the agent can always perfectly align with the teacher from any initialisation (the student flows to $\rho=1$ at $Q=\infty$). This is analogous to the student being in the easy phase at \textit{lower green region of the plot} in \cref{fig4}b, as with probability 1 the algorithm naturally converges to the optimal $\rho=1$. \Cref{suppfig2}b shows the flow for $\eta_2 = 0.05$; in this regime we observe the flow to some suboptimal $\rho$ at $Q=\infty$; this is analogous to the student being in the easy phase at \textit{the top of the plot} in \cref{fig4}b, as with probability 1 the algorithm converges to a value of $\rho$ from any initialisation. However, this value of $\rho$ is suboptimal. \Cref{suppfig2}c shows the flow for $\eta = 0.045$, we see that depending on the initial $\rho$, the agent will flow to one of two fixed point in $\rho$ at $Q=\infty$; this is analogous to the agent being in the \textit{hybrid-hard} phase in \cref{fig4}b, where with high probability the agent converges to the worse $\rho$. The `good easy phase', characterising the behaviour seen in \cref{suppfig2}a, is indicated by the green region in \cref{suppfig2}d.

\looseness=-1
\section{Phase portrait construction}\label{app:phase}

In the spherical case (constant $Q(\alpha) = Q$) with $\mathbb{I}(\Phi) = \prod_t^T\theta(y_ty^*_t)$, the ODE governing the evolution of normalised overlap $\rho$ is 

\begin{eqnarray}
    \frac{d\rho}{d\alpha} &&= \frac{\eta_1 + \eta_2}{\sqrt{2\pi Q}} \left(1-\frac{1}{\pi}cos^{-1}\left(\rho\right)\right)^{T-1} \biggl[ 1- \rho^2 \nonumber \\
    &&- \frac{\eta_1 - \eta_2}{T} \sqrt{\frac{\pi}{2}}\frac{\rho}{\sqrt{Q}}  \left(1-\frac{1}{\pi}cos^{-1}\left(\rho\right)\right)\biggr] \nonumber \\
    &&- \frac{\eta_2^2}{2T}\frac{\rho}{Q}
    \label{eq:fixedpoints}
\end{eqnarray}

The fixed points of this equation for $Q=1$ were found by numerically solving for $\frac{d\rho}{d\alpha} = 0$. We observe there are always 1 or 3 fixed points. From observing the sign of $\frac{d\rho}{d\alpha}$ either side of the fixed points, we observe that if there is one fixed point it is stable, and if there are three fixed points the outermost are stable, but the innermost fixed point sandwiched between the 1st and 3rd, is unstable. We then construct the phase plots in \ref{fig4} by sweeping over $(\eta_1, \eta_2)$ values and counting the number of fixed points -- the yellow hybrid-hard region then corresponds to a region with 3 fixed points of \ref{eq:fixedpoints}.

\section{Additional experimental details}
\label{app:exper}
\renewcommand{\thefigure}{F\arabic{figure}}
\setcounter{figure}{0}

Instructions for running simulations and experiments can be found in \href{https://github.com/nishp99/RL_Perceptron}. In both Bossfight and Pong experiments the loss was augmented to have an entropy regularisation term weighted by $\beta=0.01$ to prevent early convergence to a deterministic policy. $\beta$ is multiplied by a factor of 0.995 after each episode to gradually decrease the regularisation term contribution.

\textbf{Bossfight:} The policy network was optimised with Adam optimiser with a learning rate of $2*10^{-3}$. We discounted rewards with a discount factor of $\gamma=0.999$. The calculated returns ($G_t$) over an episode were normalised by centering around the episode mean and dividing by episode standard deviation (a standard practice for stability). 10 parallel environments (all using the same policy network for actions) were used to collect trajectories for a mean update. Generalisation performance was calculated at pre-set intervals by taking the mean of 1000 environments running in parallel (all acting on the same policy at the same time-point in training). Learning curves for 10 separately trained agents were obtained in order to calculate the mean and standard deviation which was plotted in \cref{fig6}a.

\textbf{Pong:} For enhanced stochasticity to prevent memorisation of the game, random initialisation were implemented. Where before each episode one of ten pretrained agents were randomly chosen, then the environment was ran for a random number of timesteps (randomly sampled between 10 and 55 inclusive) while acting under the policies of the pretrained agents in order to provide random initialisations. We also implemented frameskip, where for every action taken the action would be taken a random number of times (sampled between 1 and 5 inclusive). The policy network was 
optimised with Adam optimiser with a learning rate of $2*10^{-3}$. We discounted rewards with a discount factor of $\gamma=1$ (i.e no discount). The returns ($G_t$) over an episode were normalised by centering around the episode mean and dividing by episode standard deviation. 16 parallel environments (all using the same policy network for actions) were used to collect trajectories for a mean update. Generalisation performance was calculated at preset intervals by
taking the mean performance of 8 environments running in parallel (all acting on the same policy at the same time-point in training), then repeating this 20 times and taking the mean over the 20 repeats. Learning curves for 20 separately trained agents were obtained in order to calculate the mean and standard deviation which was plotted in \cref{fig:8 sa_steps}. 
\looseness=-1
\begin{table}[h!]
\caption{\label{tab:table2}
\rebuttal{\textbf{Difference of generalisation performance of Pong agents} The results of a t-test to test the significance of the difference of the generalisation performance of agents trained to play pong with reward conditioned on surviving for different times $T$}}
\begin{ruledtabular}
\begin{tabular}{ccc}
 Pairing of $T$s the test is between & Step no. & $p$-value \\
\hline
30 and 50 & $4.60\times10^6$ & 0.00096\\
30 and 70 & $4.60\times10^6$ & 0.00002\\
50 and 70 & $8.04\times10^6$ & 0.07859\\
\end{tabular}
\end{ruledtabular}
\end{table}

\nocite{*}

\cleardoublepage
\bibliography{apssamp}% Produces the bibliography via BibTeX.

\end{document}